\documentclass[Royal,sageh,times]{sagej}

\usepackage{moreverb,url}

\usepackage[colorlinks,bookmarksopen,bookmarksnumbered,citecolor=red,urlcolor=red]{hyperref}

\usepackage{pdfpages}
\usepackage{svg}
\usepackage{subcaption}
\usepackage{graphicx}
\usepackage{fontawesome}
\usepackage{booktabs}
\usepackage{amsmath}
\usepackage{url}
\usepackage{enumitem}
\usepackage{wrapfig}
\usepackage{lscape} 
\usepackage{xcolor,lipsum}
\usepackage{tcolorbox}

\usepackage{array}
\usepackage{caption} 
\usepackage[singlelinecheck=false]{caption}

\newcommand\BibTeX{{\rmfamily B\kern-.05em \textsc{i\kern-.025em b}\kern-.08em
T\kern-.1667em\lower.7ex\hbox{E}\kern-.125emX}}

\def\volumeyear{2025}

\begin{document}


\runninghead{Hamilton and Ali}

\title{Neuro-symbolic AI for Predictive Maintenance (PdM) - review and recommendations.}

\author{Kyle Hamilton\affilnum{1} and Muhammad Intizar Ali\affilnum{2}}

\affiliation{\affilnum{1}Dublin City University, Ireland\\
\affilnum{2}Dublin City University, Ireland}

\corrauth{Kyle Hamilton, Dublin City University, DCU Glasnevin Campus,
Dublin 9,
Ireland}

\email{kyle.hamilton@dcu.ie}

\begin{abstract}
In this document we perform a systematic review of the State-of-the-art in Predictive Maintenance (PdM) over the last five years in industrial settings such as commercial buildings, pharmaceutical facilities, or semi-conductor manufacturing. In general, data-driven methods such as those based on deep learning, exhibit higher accuracy than traditional knowledge-based systems. These systems however, are not without significant limitations. The need for large labeled data sets, a lack of generalizability to new environments (out-of-distribution generalization), and a lack of transparency at inference time are some of the obstacles to adoption in real world environments. In contrast, traditional approaches based on domain expertise in the form of rules, logic or first principles suffer from poor accuracy, many false positives and a need for ongoing expert supervision and manual tuning. While the majority of approaches in recent literature utilize some form of data-driven architecture, there are hybrid systems which also take into account domain specific knowledge. Such hybrid systems have the potential to overcome the weaknesses of either approach on its own while preserving their strengths. We propose taking the hybrid approach even further and integrating deep learning with symbolic logic, or Neuro-symbolic AI, to create more accurate, explainable, interpretable, and robust systems. We describe several neuro-symbolic architectures and examine their strengths and limitations within the PdM domain. We focus specifically on methods which involve the use of sensor data and manually crafted rules as inputs by describing concrete NeSy architectures. In short, this survey outlines the context of modern maintenance, defines key concepts, establishes a generalized framework, reviews current modeling approaches and challenges, and introduces the proposed focus on Neuro-symbolic AI (NESY).

\end{abstract}


\maketitle

\section{Introduction}

 






In modern industry, maintenance is a critical strategic function, representing a significant portion of operational expenditures\vphantom{, estimated to be between 15\% and 60\% of total operating costs}. The evolution towards Industry 4.0—the fourth industrial revolution characterized by the integration of Cyber-Physical Systems (CPS), the Internet of Things (IoT), and Big Data analytics—has created an opportunity to transform maintenance from a cost center into a strategic value driver. This new industrial paradigm provides the technological foundation to shift away from traditional, reactive maintenance philosophies toward more intelligent, proactive strategies.

\begin{figure}[ht]
    \centering
    \includegraphics[width=1\linewidth]{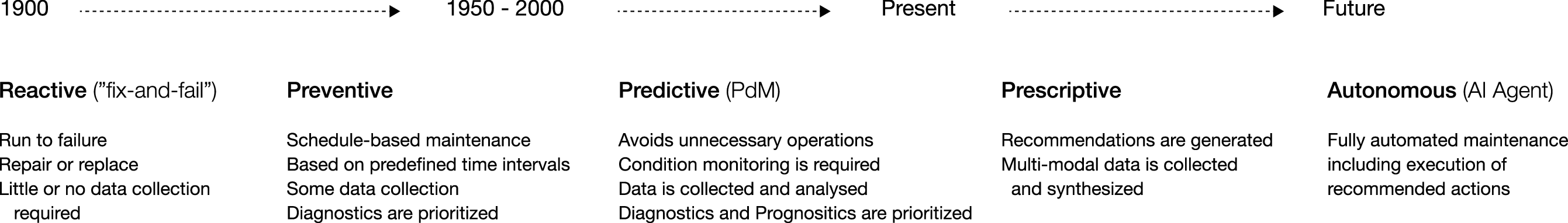}
    \caption{Maintenance implementation types timeline. Adapted from \cite{Achouch_2022}.}
    \label{fig:PdM-timeline-horz}
\end{figure}

Historically, the dominant approach to maintenance has been reactive, often described as a ``fail-and-fix'' model, where action is taken only after a component has failed. This method inevitably leads to unplanned downtime, production losses, and higher repair costs. In contrast, modern proactive approaches aim to anticipate failures before they occur \cite{Achouch_2022}. Predictive Maintenance (PdM) stands at the forefront of this transformation. By leveraging real-time data and advanced analytics, PdM aims to forecast equipment failures, enabling maintenance to be scheduled precisely when needed. The primary benefits of this strategy include improved productivity, minimized unplanned downtime, reduced system faults, and more efficient use of resources. A diagram showing the evolution of maintenance strategies is provided in Figure \ref{fig:PdM-timeline-horz}. 

A successful PdM program benefits from a standardized workflow, or framework, that guides the process from data collection to actionable insights. A generalized PdM lifecycle includes five key stages: \textit{1. Data Acquisition and Processing}, \textit{2. Health Indicator (HI) Construction and Feature Engineering}, \textit{3. Health Stage (HS) Division and Fault Detection}, \textit{4. Prognostics and Remaining Useful Life (RUL) Prediction}, and \textit{5. Decision Support and Action}. The effectiveness of the PdM framework depends heavily on the robustness and accuracy of the models used in the \textit{Detection} and \textit{Prognostics} stages. These models can be broadly categorized based on their reliance on physics (first principles), expert knowledge (Knowledge-based), or operational data (data-driven).


While data-driven methods, such as those based on deep learning, generally exhibit higher accuracy than purely knowledge-based systems, they suffer from significant limitations, including the need for large labeled data sets, poor generalization to new environments (out-of-distribution generalization), and a lack of transparency during inference. Traditional knowledge-based approaches, in turn, often suffer from poor accuracy, many false positives, and the need for ongoing expert supervision.

Hybrid systems have the potential to overcome the weaknesses of individual approaches while preserving their strengths. The reviewed literature confirms a clear trend toward hybrid and multi-model approaches, recognizing that combining strengths is necessary for modern industrial systems. We propose taking this hybrid approach further by integrating deep learning with symbolic logic, known as Neuro-symbolic AI (NESY), which leverage not only the combination of multiple deep learning models, but integrate expert knowledge and first principles to create more accurate, explainable, interpretable and trust-worthy systems. 

NESY supports trust and model validation by embedding explicit rules and logical constraints into learning systems, so stakeholders can see what the system outputs and check it against domain standards or physical laws. Because these constraints are machine-readable, they can be used as test oracles: violations highlight concrete bugs or mismatches between data-driven behavior and expert knowledge, making debugging and refinement more systematic. For regulatory, safety, and contractual requirements, neuro-symbolic models can provide traceable rationales, showing which rules, facts, and sensor patterns led to a decision, supporting audits, documentation, and certification. Finally, by encoding learned regularities in symbolic form (rules, predicates, constraints) or aligning them with an existing ontology or physics-based model, NESY turns black-box patterns into reusable, inspectable knowledge that can be maintained by engineers, combined with existing procedures, and carried across systems and projects.

Although most surveys point to hybrid solutions as the future of industrial maintenance automation, this work goes further by presenting concrete NESY architectures for PdM that integrate time-series data with rules and logic through joint learning and reasoning. We outline multiple integration strategies to offer practical research directions under common industrial constraints, such as limited data availability, the need for interpretability, and for the reduction of false positives.

The remainder of this manuscript is structured as follows. Section \ref{sec:lit-review} reviews recent PdM surveys and highlights the main research gaps. Section \ref{sec:methodology} describes our methodology, including search criteria and the study selection procedure. We present a data synthesis with an overview and descriptive statistics of the selected studies. Section \ref{sec:modeling} examines modeling approaches, covering physics-based, rule-based, data-driven, and hybrid methods. Section \ref{sec:neuro-symbolic} introduces neuro-symbolic approaches, with emphasis on compiled schemes, discusses our findings and outlines future directions. Section \ref{sec:conclusion} concludes the manuscript.


\section{Literature Review} \label{sec:lit-review}
This section synthesizes the findings from a number of recent surveys \cite{Jimenez_2020, Achouch_2022, Cheng_2019, Dalzochio_2020, Granderson_2017, Lee_Kim_2018, Lei_Li_2018, Rosato_El_Youssef_2022, Sayyad_Kumar_2021, Shi_OBrien_2019, Silvestri_2020, Ul_Hassan_2025, Vrignat_Kratz_Avila_2022, Zonta_Costa_2020, Ignjatovska_Pandilov_Petreski_2023, Mallioris_Aivazidou_Bechtsis_2024, 8707108, Psarommatis_May_Azamfirei_2023, Lu_Afridi_Kang_Ruchkin_Zheng_2024, Fitas, Li_Liu_Sun_Qin_Chu_2024,Wilhelm_2021} describing the State-of-the-art in PdM. Among these, several focus specifically on hybrid \cite{Jimenez_2020, Dalzochio_2020, Wilhelm_2021, Fitas}, and neuro-symbolic \cite{Lu_Afridi_Kang_Ruchkin_Zheng_2024} approaches. This section also serves as an introduction to the subject by outlining key concepts, providing a generalized framework for predictive maintenance, and describing the typology of current modeling approaches. Future directions, as found in the literature, are also listed. Finally a case is given for neuro-symbolic approaches for the future of PdM: fully autonomous predictive maintenance.

\subsection{Predictive Maintenance Definition and Key Concepts} \label{sec:pdm-definition}

To navigate the landscape of modern maintenance, it is essential to clarify the relationship between several key concepts prevalent in the literature:
\begin{itemize}
    \item \textbf{Predictive Maintenance (PdM)}: This is the core strategy of forecasting the future health of an asset (e.g., Air Handling Unit, or AHU) to determine the optimal time for maintenance.
    \item \textbf{Condition-Based Maintenance (CBM)}: A key tactic within the broader PdM strategy, CBM operationalizes prediction by recommending maintenance actions based on real-time information from condition monitoring, triggering interventions when specific thresholds are crossed rather than relying on a long-term forecast alone.
    \item \textbf{Prognostics and Health Management (PHM)}: This is a comprehensive management cycle that encompasses three key functions: observation (monitoring system health), analysis (diagnosing faults and predicting future degradation), and action (making informed maintenance decisions).
    \item \textbf{Prescriptive Maintenance}: By analyzing historical records together with real-time data, the system forecasts needed maintenance and recommends specific actions. Prescriptive maintenance shifts operations from scheduled preventive work to proactive, intelligent maintenance planning \cite{Matyas_2017}. 
    \item \textbf{Remaining Useful Life (RUL)}: A critical output of the prognostics process, RUL is defined as the estimated time an asset is likely to operate before it requires repair or replacement.
\end{itemize}

\subsubsection{A Generalized Framework for Predictive Maintenance}\label{sec:pdm-framework}
The implementation of a successful Predictive Maintenance program benefits from a standardized workflow that guides the process from data collection to actionable insights. By synthesizing various process models from the academic literature \cite{Achouch_2022, Sayyad_Kumar_2021}, we can establish a single, cohesive framework that outlines the key stages of a generalized PdM lifecycle.
\begin{enumerate}
    \item \textbf{Data Acquisition and Processing}: This initial stage involves capturing and storing monitoring data from a wide array of sources. These include physical sensors such as accelerometers, acoustic emission sensors, and thermometers, as well as operational data from Computerized Maintenance Management Systems (CMMS) and event logs. A significant challenge in this phase is acquiring sufficient high-quality, run-to-failure data, as equipment in real-world settings is rarely allowed to operate until complete failure. This scarcity has led to the widespread use of public datasets, such as the NASA Turbofan engine degradation dataset and the FEMTO bearing dataset, for developing and validating PdM models. The fidelity of the entire PdM process is fundamentally constrained by the quality and relevance of the data captured in this foundational stage. A list of popular public datasets is provided in Appendix \ref{app:datasets}.
    \item \textbf{Health Indicator (HI) Construction and Feature Engineering}: Raw sensor data is often noisy and too complex for direct use in predictive models. Therefore, this stage focuses on processing the raw data to construct features, or Health Indicators (HIs), that more clearly represent the health status and degradation trend of the system. HIs can be categorized as:
    \begin{itemize}
        \item \textbf{Physics-based HIs (PHIs)}: These are derived from statistical methods or signal processing and have a direct physical meaning (e.g., Root Mean Square (RMS), kurtosis).
        \item \textbf{Virtual HIs (VHIs)}: These are constructed by fusing multiple signals or features using dimensionality reduction techniques like Principal Component Analysis (PCA), providing a composite view of system health. The quality of the constructed HI is paramount, as it directly determines the signal-to-noise ratio for the subsequent prognostic models; a poorly constructed HI can obscure the degradation trend, rendering even the most sophisticated RUL prediction algorithm ineffective.
    \end{itemize}
    \item \textbf{Health Stage (HS) Division and Fault Detection}: The purpose of this stage is to identify when a system deviates from its normal operating condition. This process generates symptoms that are crucial for subsequent diagnosis. The division can be a simple two-stage process, where the goal is to detect an incipient fault to determine a First Predicting Time (FPT), which triggers the RUL prediction. For systems with more complex degradation patterns, a multi-stage division may be necessary to characterize different phases of deterioration (e.g., healthy, moderate degradation, severe degradation). An accurately determined FPT is critical for resource efficiency, preventing the commitment of computational resources to prognostics before a meaningful degradation trend has even begun.
    \item \textbf{Prognostics and RUL Prediction}: This is the core predictive task of the framework. Using the HIs and the identified health stages, this stage focuses on modeling the degradation trend to forecast the RUL of the equipment. Various modeling techniques are employed to project the current health state into the future, predicting the point at which the equipment will reach its end-of-life or a predefined failure threshold. This stage represents the analytical core of the PdM strategy, where historical patterns are transformed into forward-looking intelligence on asset longevity.
    \item \textbf{Decision Support and Action}: The final stage translates the outputs of the diagnostic and prognostic models into actionable maintenance recommendations. This decision-making process involves scheduling maintenance interventions by optimizing for a range of factors, including maintenance costs, resource availability, spare parts inventory, and the impact on production schedules. The ultimate goal is to seamlessly integrate the predictive insights into the broader operational workflow of the organization. This final stage is where the technical outputs of PdM are translated into tangible business value, bridging the gap between advanced analytics and operational reality.
\end{enumerate} 

The effectiveness of this framework depends heavily on the robustness and accuracy of the models used in the fault detection and prognostics stages. The following section provides a typology of the primary modeling approaches employed in modern PdM.

\subsection{Typology of Predictive Maintenance Modeling Approaches:}\label{sec:pdm-typology}
PdM modeling approaches can be broadly categorized based on their fundamental reliance on either first principles of physics, expert knowledge, or historical operational data. Each category offers distinct advantages and disadvantages, making the choice of model dependent on the specific application, data availability, and system complexity \cite{Sayyad_Kumar_2021, Zonta_Costa_2020, Silvestri_2020}.

\subsubsection{Physics-Based (Model-Driven) Approaches:}
This approach relies on building mathematical models from the first principles of damage and failure mechanisms. These models use deep, explicit knowledge of a system's physical behavior, such as material properties and stress levels, to describe its degradation process. Examples include thermodynamics or vibration analysis. By understanding the underlying physics of failure, these models can offer highly accurate and interpretable predictions. The strengths and weaknesses of these models are summarized in Table \ref{tab:sw-physics}.

\begin{table}
    \centering
    \footnotesize
    \renewcommand{\arraystretch}{1.5}
    \begin{tabular}[b]{p{5.75cm}p{5.75cm}}
        \hline
         \textbf{Strengths} & \textbf{Weaknesses}\\
         \hline
         Potential for high accuracy and precision. & Difficult and costly to develop for complex systems.\\
         Highly explainable, as predictions are tied to physics. & Requires deep domain expertise and knowledge of failure modes. \\
         Does not require large amounts of historical data. & May not account for unexpected failure modes or conditions.\\
         \hline
    \end{tabular}
    \caption{The strengths and weaknesses of Physics-Based approaches.}
    \label{tab:sw-physics}
\end{table}

\subsubsection{Data-Driven (Process History-Based) Approaches:}
Data-driven approaches build predictive models by identifying patterns and correlations directly from large volumes of historical operational and failure data. These methods do not require an explicit mathematical model of the system's physics, making them highly versatile for complex systems where physical modeling is impractical. This is the most prevalent category in recent PdM literature, fueled by advances in sensor technology and machine learning. These models can be further categorized:
\begin{itemize}
    \item \textbf{Statistical Models}: These techniques use statistical principles to model degradation and predict failure. Common methods include Autoregressive (AR) models, stochastic process models (Wiener, Gamma, and Inverse Gaussian processes), and Proportional Hazards Models (PHM), which integrate event data with condition monitoring data.
    \item \textbf{Machine Learning (ML) Models}: These models are adept at learning complex, non-linear relationships from data. Widely used techniques include Artificial Neural Networks (ANNs) and their variants like Recurrent Neural Networks (RNNs) and Long Short-Term Memory (LSTM) networks, Support Vector Machines (SVM), Random Forests (RF), and Gaussian Process Regression (GPR).
    \item \textbf{Deep Learning (DL) Models}: A subset of machine learning, deep learning uses advanced neural network architectures to automatically learn features from raw data. Models like Convolutional Neural Networks (CNNs) and Auto-encoders (AEs) are increasingly used for both feature extraction and RUL prediction.
\end{itemize}

The strengths and weaknesses of Data-Driven approaches are summarized in Table \ref{tab:sw-datadriven}.
\begin{table}
    \centering
    \footnotesize
    \renewcommand{\arraystretch}{1.5}
    \begin{tabular}[b]{p{5.75cm}p{5.75cm}}
        \hline
         \textbf{Strengths} & \textbf{Weaknesses}\\
         \hline
         Can model complex systems without deep physical knowledge. & Requires large amounts of high-quality, labeled training data, making them susceptible to the ``Data Scarcity and Quality'' challenge.\\
         Highly effective when large datasets (Big Data) are available. & Risk of overfitting, where the model performs poorly on new data due to factors such as variability of machine operating conditions. \\
         Can identify subtle patterns missed by human experts. & Often a ``black-box'' with low explainability, a barrier driving the need for Explainable AI (XAI) as a key future direction.\\
         \hline
    \end{tabular}
    \caption{The strengths and weaknesses of Data-Driven (Process History-Based) approaches.}
    \label{tab:sw-datadriven}
\end{table}

\subsubsection{Knowledge-Based Approaches:}
This category of models relies on capturing and codifying knowledge from human experts to create a basis for reasoning about system health. The knowledge is typically structured to guide diagnostic and prognostic tasks. The primary types of knowledge-based models include:
\begin{itemize}
    \item \textbf{Rule-Based Systems}: These use a set of ``IF-THEN'' rules to represent expert knowledge and make decisions.
    \item \textbf{Case-Based Reasoning}: This method solves new problems by retrieving and adapting solutions from a database of similar past cases.
    \item \textbf{Fuzzy Logic Systems}: This approach is particularly useful for managing uncertainty and imprecision by using terms such as ``high'' or ``low'', rather than precise numerical values.
    \item \textbf{Ontologies and Reasoning}: An ontology is a graph structure which enriches information with meta-data and/or logical constraints, and enables automated reasoning. As an example, an expert may formalize an ontology to support vibration analysis of machine components. That ontology can then be leveraged to define SWRL\endnote{\url{https://www.w3.org/submissions/SWRL/}} rules that infer potential failures.
\end{itemize}

The strengths and weaknesses of Knowledge-Based approaches are summarized in Table \ref{tab:sw-knowledge}.
\begin{table}
    \centering
    \footnotesize
    \renewcommand{\arraystretch}{1.5}
    \begin{tabular}[b]{p{5.75cm}p{5.75cm}}
        \hline
         \textbf{Strengths} & \textbf{Weaknesses}\\
         \hline
         Can effectively incorporate expert and domain knowledge. & Knowledge acquisition can be a difficult and biased process. \\
         Highly explainable and transparent in its reasoning. &  Limited applicability for complex prognostic tasks.\\
         Can operate with incomplete or uncertain information. & Difficult to maintain and update the knowledge base.\\
         \hline
    \end{tabular}
    \caption{The strengths and weaknesses of Knowledge-Based approaches.}
    \label{tab:sw-knowledge}
\end{table}

\subsubsection{Hybrid and Multi-Model Approaches:}
Hybrid approaches combine two or more models from the categories above to leverage their respective strengths while mitigating their weaknesses. This strategy is gaining traction as a way to achieve higher accuracy and robustness, particularly for complex industrial systems. For example, a physics-based model could be used to describe the general degradation trend, while a data-driven model refines the prediction using real-time sensor data. Similarly, a knowledge-based system could be used to fuse the outputs of multiple data-driven models. The strengths and weaknesses of Hybrid and Multi-Model approaches are summarized in Table \ref{tab:sw-hybrid}.
\begin{table}
    \centering
    \footnotesize
    \renewcommand{\arraystretch}{1.5}
    \begin{tabular}[b]{p{5.75cm}p{5.75cm}}
        \hline
         \textbf{Strengths} & \textbf{Weaknesses}\\
         \hline
         Potential for higher accuracy and robustness. & Increased complexity in design and implementation. \\
         Ability to handle multiple modalities (eg., text, images, signals). & Requires expertise across multiple modeling domains. \\
         More flexible and adaptable to different problems. & Difficult to integrate and fuse different model outputs. \\
         \hline
    \end{tabular}
    \caption{The strengths and weaknesses of Hybrid and Multi-Model approaches.}
    \label{tab:sw-hybrid}
\end{table}

\subsection{Prevailing Challenges in Predictive Maintenance Implementation}
While the promise of Predictive Maintenance is significant, its widespread adoption is impeded by a series of practical challenges that span the entire implementation lifecycle \cite{Jimenez_2020, Dalzochio_2020, Wilhelm_2021}. These challenges directly influence the selection and viability of the modeling approaches discussed previously. For instance, pervasive data scarcity often limits the practical application of purely data-driven methods and motivates the use of physics-based or hybrid alternatives that are less reliant on extensive historical datasets. The challenges across data, modeling, and operations are outlined below.

\subsubsection{Data-Related Challenges}
\begin{itemize}
    \item \textbf{Data Scarcity and Class Imbalance}: The most effective data-driven models require large volumes of high-quality, labeled run-to-failure data. However, this type of data is notoriously difficult and expensive to obtain in industrial settings, where equipment is typically repaired before catastrophic failure. This scarcity of authentic run-to-failure data poses the single greatest threat to the efficacy of the generalized PdM framework, particularly for the data-hungry modeling techniques used in Stage 4 (Prognostics and RUL Prediction). This class imbalance often forces researchers and practitioners to rely on simulation benches or limited public datasets.
    
    \item \textbf{Data Heterogeneity and Fusion}: Modern industrial systems generate data from a wide variety of sources, including time-series sensor data, text-based maintenance logs, and structured design information from Building Information Modeling (BIM). Integrating and fusing these diverse and heterogeneous data streams into a cohesive format suitable for analysis is a challenge.
    
    \item \textbf{Data Standards and Infrastructure}: The lack of common standards for data tagging and semantic representation (though efforts like Project Haystack\endnote{\url{https://project-haystack.org/}} and Brick Schema\endnote{\url{https://brickschema.org/}} are emerging) creates significant barriers to interoperability and scalability. Furthermore, organizations often face challenges with IT integration, network bandwidth and latency for real-time monitoring, and ensuring robust data security against cyber threats.
\end{itemize}

\subsubsection{Model-Related Challenges}
\begin{itemize}
    \item \textbf{Model Accuracy and Generalization}: There is a notable lack of a universal model that can be applied across different scenarios. Models developed and validated under controlled laboratory conditions often fail to generalize to complex, real-world systems that operate under varying and dynamic conditions. Extrapolating model performance from one asset or context to another remains a significant challenge. 
    \item \textbf{Uncertainty Management}: RUL prediction is inherently uncertain due to the stochastic nature of degradation processes and the influence of time-varying operational and environmental conditions. Developing models that can effectively quantify and manage this uncertainty is essential for making reliable maintenance decisions.
    \item \textbf{Overfitting and Training}: Complex machine learning models are susceptible to overfitting, where they learn the noise in the training data rather than the underlying patterns, leading to poor performance on new data. This requires robust training and validation techniques, such as the use of customized loss functions that penalize false positives or false negatives differently, and strategies to handle the class imbalance inherent in failure data.

\end{itemize}

\subsubsection{Operational and Organizational Challenges}
\begin{itemize}
    \item \textbf{Cost and Return on Investment (ROI)}: The initial investment for PdM can be substantial, including the cost of sensors, data infrastructure, software, and specialized expertise. Articulating a clear and convincing ROI is often a major hurdle for securing organizational buy-in and funding.
    \item \textbf{Multidisciplinary Expertise}: Effective PdM implementation requires a rare combination of skills, blending deep domain knowledge in engineering with expertise in data science, IT, and software development. Few organizations possess this multidisciplinary talent in-house, making reliance on external consultants or service providers common.
    \item \textbf{Integration and Human Factors}: Integrating PdM outputs with existing enterprise systems, such as production scheduling and CMMS, is a critical but often overlooked challenge. Furthermore, current industrial machines generally lack self-maintenance capabilities, meaning the process remains dependent on human operators to interpret the model outputs and execute the recommended actions.
\end{itemize}

Overcoming these challenges will require concerted efforts across research, technology development, and organizational change. The following section lists some of the opportunities for future research and development identified in the literature.

\subsection{Future directions identified in previous literature reviews}
While data-driven methods, particularly those based on machine learning, have become increasingly prevalent, there is a clear and growing trend towards hybrid and multi-model approaches. This reflects a recognition that combining the strengths of physics-based, data-driven, and knowledge-based models is often necessary to tackle the complexity of modern industrial systems. Despite significant progress, the journey towards widespread, autonomous maintenance is still fraught with challenges related to data, modeling, and operational integration. The path forward is illuminated by several key research opportunities and strategic directions.
\begin{itemize}
    \item \textbf{Real-World Validation}: A pressing need exists to move beyond theoretical frameworks and laboratory-based experiments. Implementing and validating PdM models in real industrial environments is critical to precisely evaluate their effectiveness, measure true cost savings, and build confidence among industry stakeholders.
    \item \textbf{Collaborative Datasets}: The scarcity of high-quality, labeled run-to-failure data remains a primary bottleneck. The creation of more comprehensive, publicly available datasets with verified ground-truth fault instances would be invaluable for benchmarking different PdM methodologies and accelerating innovation in the field.
    \item \textbf{Advanced Hybrid Models}: There is consensus that future research should focus on the development of more sophisticated multi-model systems capable of effectively fusing heterogeneous data sources. These models should also be able to incorporate external influences, such as environmental conditions or production schedules, for more context-aware predictions.
    \item \textbf{Explainable AI (XAI)}: The ``black-box'' nature of many advanced machine learning models is a significant barrier to their adoption, especially in safety-critical systems. Research into XAI aims to make these models more transparent and their predictions more interpretable, which is crucial for building trust and enabling human operators to make informed decisions based on the model's outputs. Interpretability is especially important when it comes to Root Cause Analysis (RCA), an active research area of industrial maintenance.
    \item \textbf{Integration and Automation}: The ultimate goal of PdM is to streamline the workflow from prediction to action. This requires tighter integration between PdM systems and enterprise-level platforms like CMMS and Enterprise Resource Planning (ERP) systems. Achieving this level of integration will pave the way for more automated decision-making and, eventually, a future of more autonomous maintenance operations.
\end{itemize}

\subsection{Literature Review Summary}
\begin{figure}[ht]
    \centering
    \includegraphics[width=.75\linewidth]{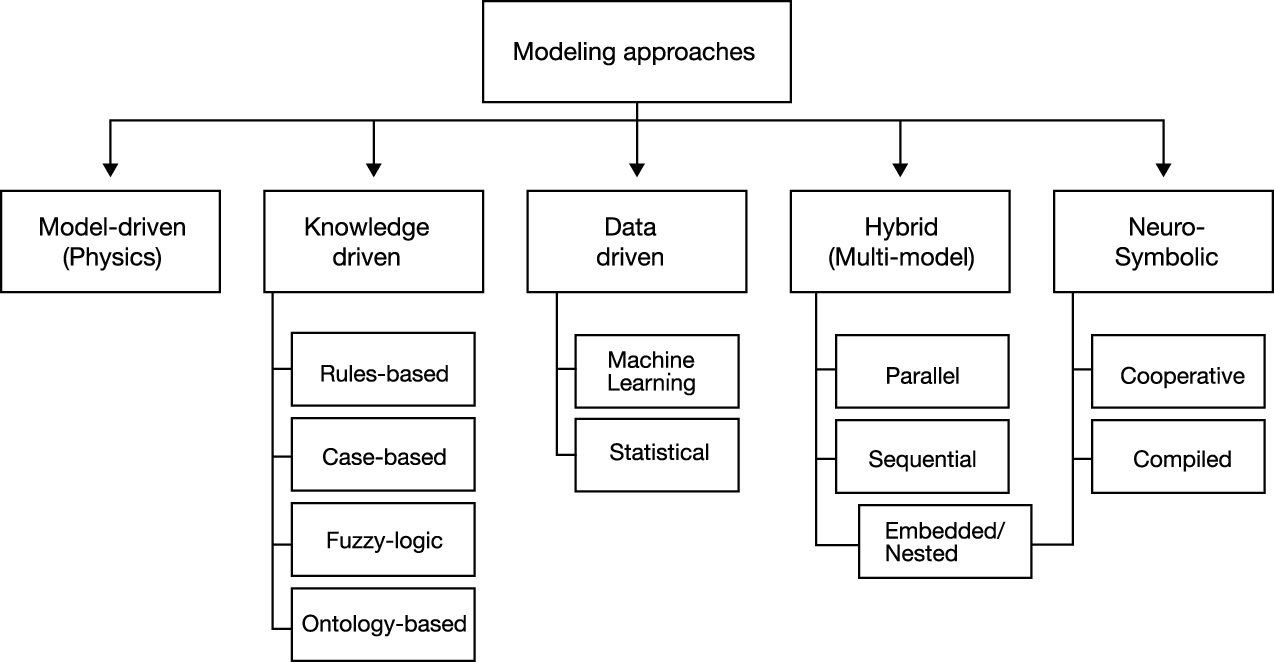}
    \caption{Typology of modeling approaches.}
    \label{fig:pdm_approaches}
\end{figure}

A synthesis of the reviewed literature reveals several overarching trends and persistent challenges. A clear thematic convergence is the emergence of ``prescriptive maintenance'' as the next frontier beyond predictive capabilities, a concept highlighted as a future direction in multiple analyses \cite{Silvestri_2020, Zonta_Costa_2020, Mallioris_Aivazidou_Bechtsis_2024}. Methodologically, there is a distinct evolution from purely physics-based models toward data-driven approaches, with a significant and growing emphasis on hybrid systems that combine the strengths of both to improve accuracy and interpretability \cite{Zonta_Costa_2020, Ignjatovska_Pandilov_Petreski_2023, Jimenez_2020}. Within hybrid systems, neuro-symbolic systems are sometimes implicitly categorized as a subset thereof. \cite{Jimenez_2020} refer to \textit{embedded structures} as one of three types of hybrid systems, but which also meet the criteria for \textit{nested} neuro-symbolic \cite{Hamilton_Nayak_Bozic_Longo_2024} (see also Section \ref{sec:neuro-symbolic}). A diagram illustrating approach types is presented in Figure \ref{fig:pdm_approaches}.

Several common challenges have also been identified. The difficulty in acquiring sufficient failure data for model training is a frequently cited impediment \cite{Mallioris_Aivazidou_Bechtsis_2024}, as is the need for more realistic, publicly available industrial datasets that move beyond laboratory conditions \cite{Lei_Li_2018}. Furthermore, the complexity of fusing heterogeneous data from disparate sources remains a significant technical hurdle \cite{Silvestri_2020, Jimenez_2020}. Finally, the scope of maintenance research is expanding beyond purely technical considerations to include non-technical factors, with a growing recognition of research gaps in sustainability and human factors, which are critical for holistic and responsible implementation in Industry 5.0 contexts \cite{Psarommatis_May_Azamfirei_2023, Silvestri_2020}.

While the majority of surveys identify hybrid solutions as a path forward in industrial maintenance automation, only one focuses on Neuro-symbolic AI for time-series data,   specifically, Artificial Intelligence of Things (AIoT) or the integration of Artificial Intelligence (AI) with the Internet of Things (IoT). The authors provide several canonical examples of NESY systems as well as describing well established advantages such as interpretability, verifiability and testability, and challenges such as the need for manually crafted functions, lack of support for multi-modal sensor data, trade-offs between interpretability and performance, and technical complexities in integrating neural networks and symbolic AI \cite{Lu_Afridi_Kang_Ruchkin_Zheng_2024}. Different from \cite{Lu_Afridi_Kang_Ruchkin_Zheng_2024}, this survey showcases concrete NESY architectures for PdM that integrate time-series data with rules and logic. We detail key forms of symbolic knowledge representation, ``a cornerstone of a neural-symbolic system'' \cite{Garcez_Gori_Lamb_Serafini_Spranger_Tran_2019}, and discuss the types of outcomes each can support. Our goal is to offer actionable research paths tailored to readers’ constraints — whether dealing with scarce data, requiring interpretability, or aiming to cut false positives, a common challenge for many PdM systems \cite{Granderson_2017}.










\section{Methodology}\label{sec:methodology}
This work followed the systematic literature review guidelines of \cite{Kitchenham07guidelinesfor}, framed around three research questions on existing and neuro‑symbolic PdM approaches and their limitations:

\begin{enumerate}
    \item What are the current State-of-the-art approaches for predictive maintenance in industrial settings?
    \item What are the limitations of current State-of-the-art approaches?
    \item How can the limitations of current approaches be addressed with neuro-symbolic techniques?
\end{enumerate}

\begin{figure}[ht]
\begin{subfigure}{0.49\textwidth}
    \centering
    \includegraphics[width=1\linewidth]{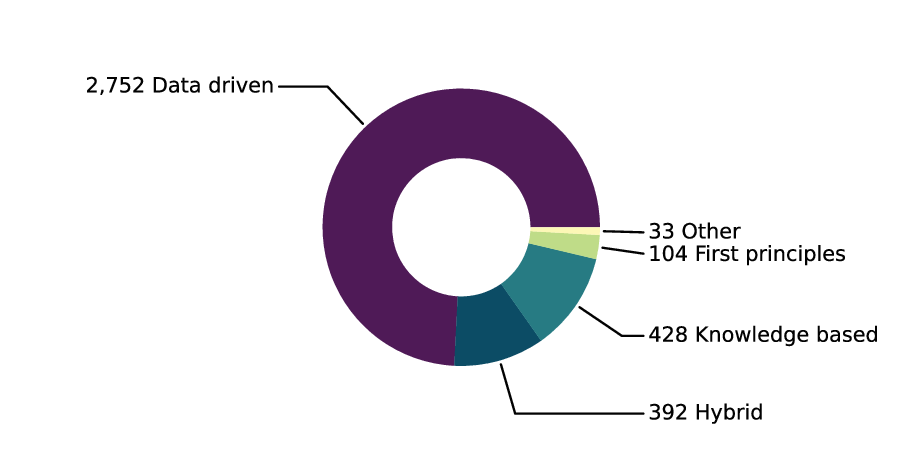}
    \caption{Number of articles in each of the modeling categories.  }
    \label{fig:Modeling_cats}
\end{subfigure}
\begin{subfigure}{0.49\textwidth}
    \centering
    \includegraphics[width=1\linewidth]{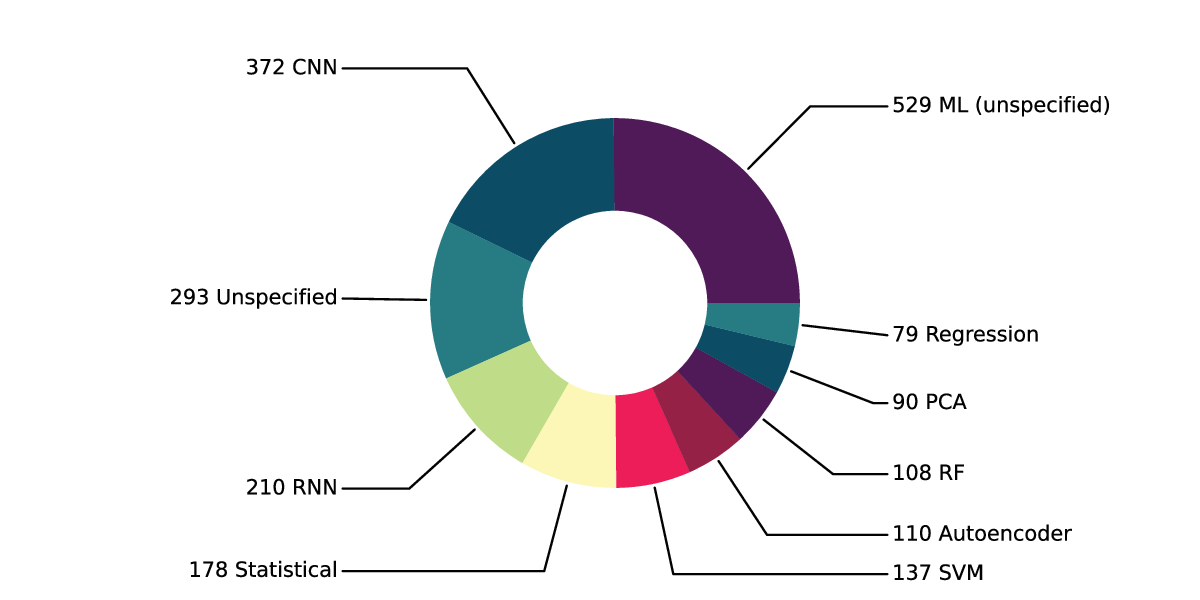}
    \caption{Number of articles in each of the top 10 data-driven modeling techniques.}
    \label{fig:dd_tech}
\end{subfigure}
\hfill
\begin{subfigure}{0.49\textwidth}
    \centering
    \includegraphics[width=1\linewidth]{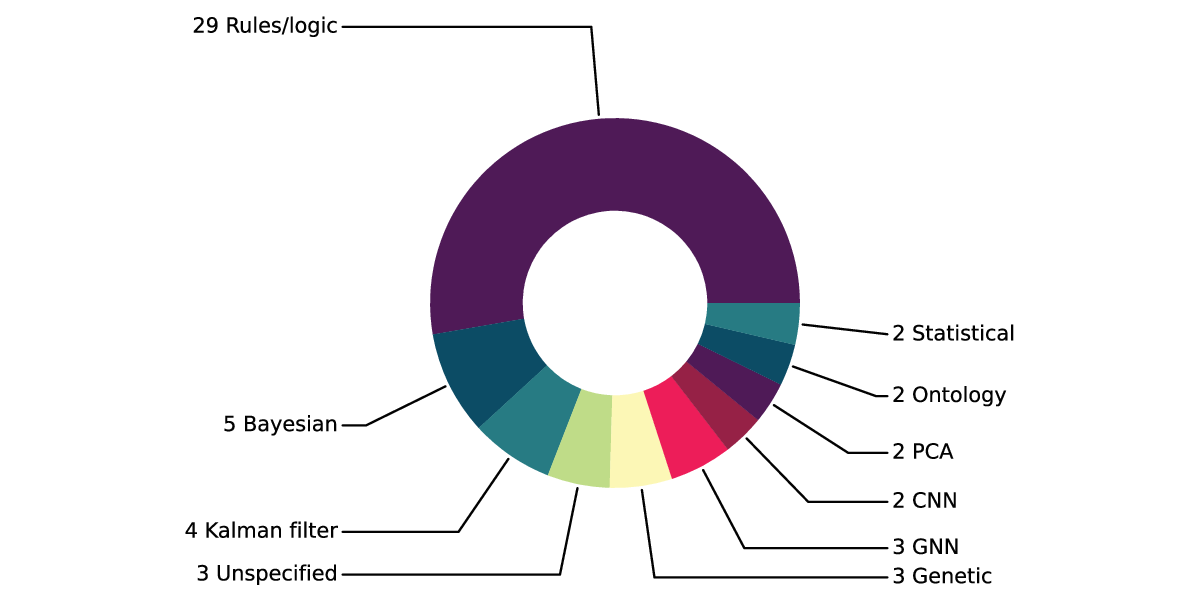}
    \caption{Number of articles in each of the top 10 knowledge-based modeling techniques.}
    \label{fig:know_tech}
\end{subfigure}
\begin{subfigure}{0.49\textwidth}
    \centering
    \includegraphics[width=1\linewidth]{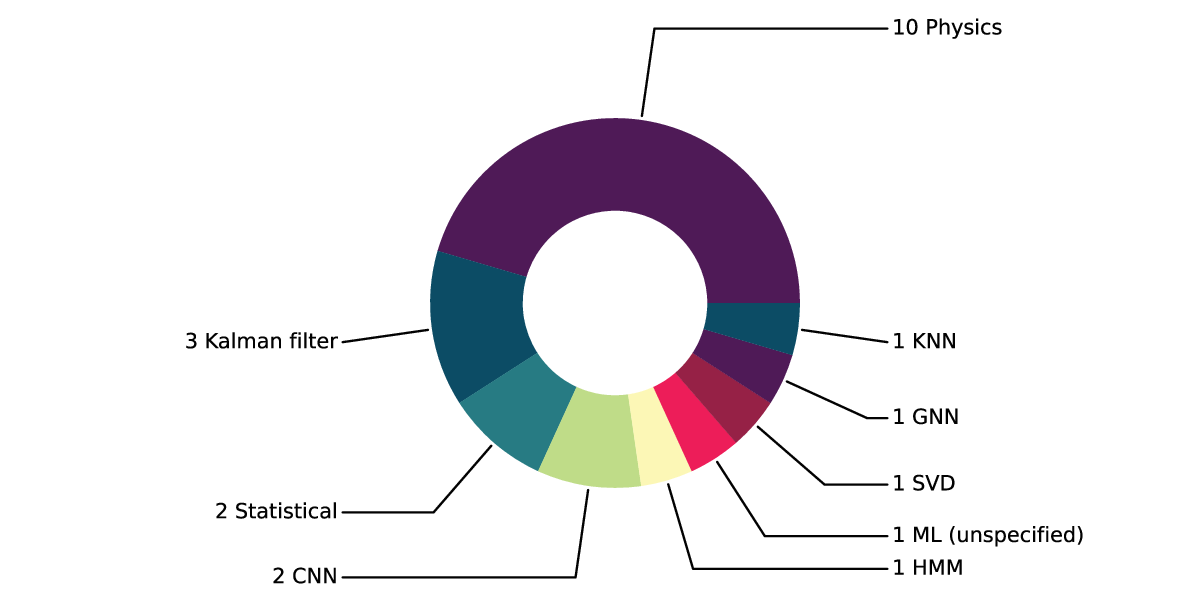}
    \caption{Number of articles in each of the top 10 physics-based modeling techniques.}
    \label{fig:first_tech}
\end{subfigure}

\caption{Modeling categories and Techniques.}
\label{fig:modeling_cat_tech}
\end{figure}

A Scopus search (query in Table \ref{tab:scopus_query}) was run in August 2025, restricted to English‐language engineering and computer science publications from 2020–2025 and excluding review papers, yielding 9,005 records. Titles and abstracts were then processed with an LLM (OpenAI gpt‑4.1‑nano) to automatically extract seven metadata fields: domain, input, output, modeling category, technique, explainability, and relevance, after which domain labels were manually cleaned and clustered with OpenRefine into 46 domain categories; non‑relevant domains (e.g., sport, audio, 3D printing) were removed, leaving 3,495 studies. Finally, descriptive statistics over these annotated records were used to characterize prevailing inputs, outputs, modeling approaches, and explainability practices in PdM.  Figure \ref{fig:Modeling_cats} shows the number of articles for each of the modeling categories. It should be noted that many of the data-driven approaches are in fact hybrid without explicitly specifying so. Figures \ref{fig:dd_tech}, \ref{fig:know_tech}, and \ref{fig:first_tech} show the breakdown of the top ten modeling techniques within the data-driven, knowledge-based, and physics-based categories, respectively. The full methodological details are provided in Appendix \ref{app:methodology}.

\section{Modeling approaches}\label{sec:modeling}
In the PdM literature, a variety of modeling approaches have been proposed, each with distinct assumptions, requirements, and strengths. This section outlines four main categories: physics-based models, which derive behavior from underlying physical laws; knowledge-based models, which encode expert rules and heuristics; data-driven models, which learn patterns directly from historical and real-time data; and hybrid models, which combine elements of the previous three to leverage complementary advantages. We provide examples from the literature and highlight trade-offs in accuracy, interpretability, and implementation effort where applicable.

\subsection{Physics-based} \label{sec:physics-based}

\begin{figure}
    \centering
    \includegraphics[width=1\linewidth]{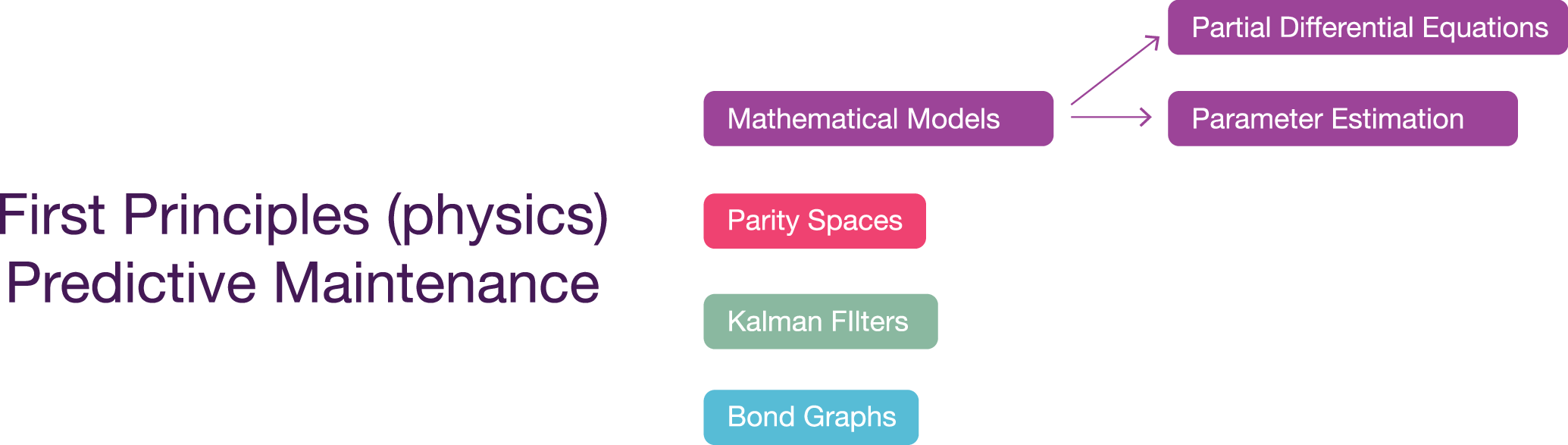}
    \caption{Common types of Physics-based/First principles Predictive Maintenance.}
    \label{fig:phys-diagram}
\end{figure}

Physics-based approaches utilize mathematical models based on first principles, or the underlying physics of the system such as thermodynamics or vibration analysis. Typical methods based on physical models for Fault Detection and Diagnosis (FDD) are bond graphs, parity equations, parameter estimation, state estimation (Kalman filter (KF)) and state observers \cite{Wilhelm_2021,Tidriri_2016}. Knowledge of the physics of a system has also been used to augment data for training NNs in situations where data is scarce, or when only healthy data exists \cite{Zgraggen_Guo_Notaristefano_Huber_2022}. However, they are less commonly used because they are difficult to generalize and require highly detailed knowledge of a system's physical characteristics. A comprehensive audit of physics based systems is beyond the scope of this survey, however, some common methods are described below - Figure \ref{fig:phys-diagram}.

\subsubsection{Kalman Filter (KF)} uses a system's dynamic model (e.g., physical laws of motion), known control inputs to that system, and multiple sequential measurements (such as from sensors) to form an estimate of the system's varying quantities (its state) that is better than the estimate obtained by using only one measurement alone. As such, it is a common sensor fusion and data fusion algorithm. KFs appeared in 74 of the studies under review. 

\subsubsection{Parity equations and bond graphs} - The parity space is a space in which all elements are residuals (or parity vectors). Residuals and parity vectors are synonymous in this context. The parity space can also be called a residual space. The relation (or equation) which generates the residual (parity vector) is called a parity relation (or equation) \cite{Patton_Chen_1991}. Parity equations are chosen to yield parity vectors (residuals) that are zero when the components are functioning perfectly, but that have a subset deviating from zero when a particular system component malfunctions. 

A Bond Graph (BG) is a graphical representation of a physical dynamic system.  A BG uses arrows (bonds) to show where energy goes, and blocks to show what parts (assets) do with that energy. BGs model systems based on the flow and exchange of power, which is represented as the product of two variables: effort (e.g., force, voltage) and flow (e.g., velocity, current):

$$
p = f \cdot e
$$


BGs yield a set of state-space equations that describe the system dynamics. These equations, or analytical redundancy relations (ARRs) \cite{Staroswiecki_2001}, should balance if the system is healthy. These are exactly the parity equations described above. In a bond-graph model, each component is parameterized by a physical feature—R (dissipation such as friction, leakage, electrical or thermal resistance), C (compliance/capacitance), I (inertia/inductance), and transducer gains (TF/GY). Parametric degradation is the gradual drift of these features from nominal values as a consequence of wear, fouling, aging, or damage. To identify faulty components, residuals are computed as differences between measured and model-predicted efforts/flows at bonds.

A significant advantage of bond graphs is their ability to easily couple multiple physical domains—such as mechanical, electrical, or thermal—in a single model. As a physics-based tool, they can also be used to generate synthetic data \cite{Keizers_2025} that accurately reflects the underlying physical principles of degradation, overcoming the limitations of real-world data availability.

However, BGs require substantial modeling effort, are sensitive to noise, and suffer from computational complexity. As such they are mainly used in hybrid, such as \cite{Borutzky_2020}, or NESY systems which mitigate these weaknesses. While both bond graphs and parity spaces were mentioned in a handful of studies, it was always in combination with data-driven methods. 



%

In brief, physics-based methods build predictive maintenance models from first principles (e.g., dynamics, thermodynamics, circuit laws), yielding interpretable parameters and residuals that can extrapolate to new operating points and work with sparse fault data, but they require detailed asset knowledge, careful calibration, and can be hard to generalize across variants. Physics-based models are strong for root-cause inference and RUL via physical parameters (e.g., stiffness, leakage, resistance).

\subsection{Knowledge-based} \label{sec:knowledge-based}
\begin{figure}
    \centering
    \includegraphics[width=1\linewidth]{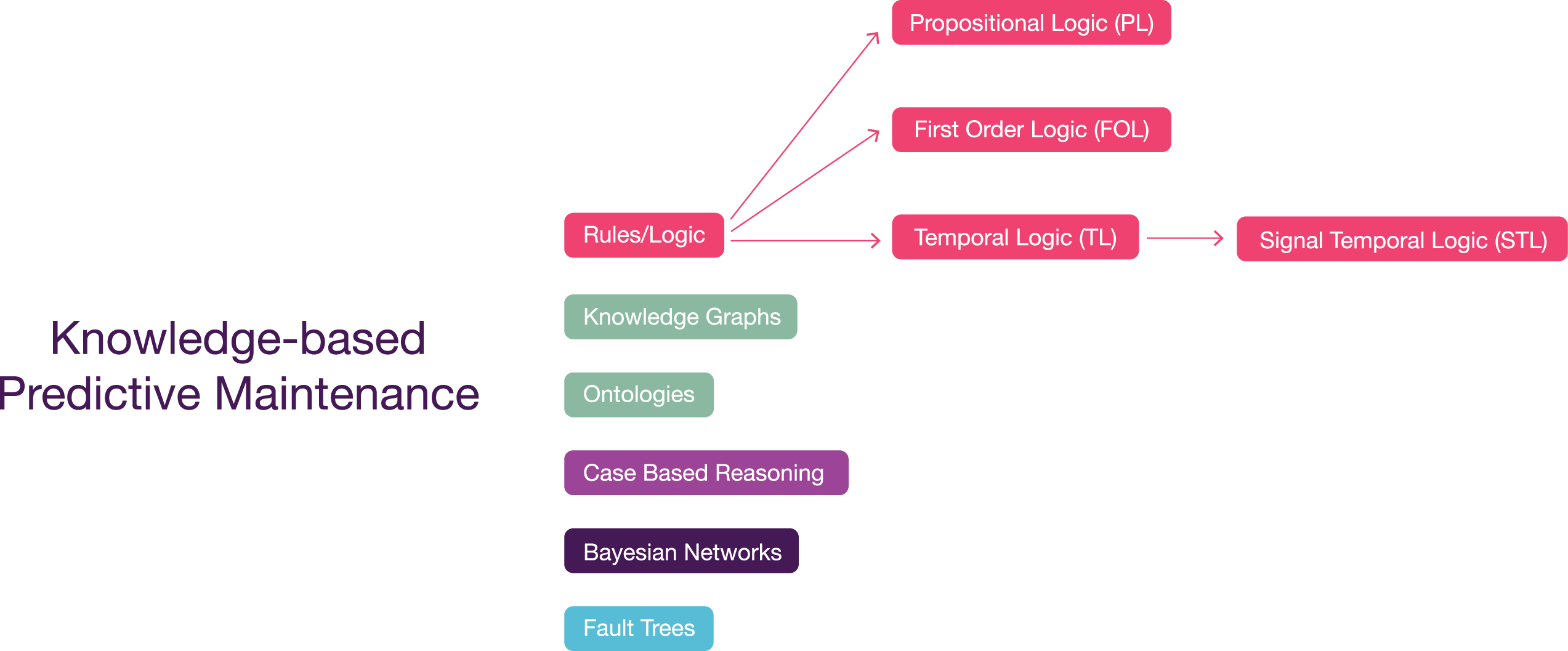}
    \caption{Types of Knowledge-based Predictive Maintenance (PdM).}
    \label{fig:KB-diagram}
\end{figure}

Knowledge-based PdM systems rely on explicit, human-understandable representations of how equipment behaves, fails, and should be serviced. They encode domain expertise through rules and logic (if–then rules and formal reasoning), ontologies (structured vocabularies capturing assets, components, and their relationships), Bayesian Networks (graphical models of probabilistic dependencies among faults, symptoms, and operating conditions), fault trees (hierarchical models of how basic events combine into system-level failures), and case-based reasoning (reusing and adapting solutions from past maintenance cases) - Figure \ref{fig:KB-diagram}. This section introduces these main forms of symbolic knowledge representation and examines how they can be used for FDD, RUL, and RCA.

\subsubsection{Logic} is the study of entailment relations—languages, truth conditions, and rules of inference. \cite{Brachman_Levesque_2004, Dyer}. A logic includes:
    \begin{itemize}
        \item \textbf{Syntax}: specifies the symbols in the language and how they can be combined to form sentences. Facts about the world are represented as \textit{sentences} in logic.
        \item \textbf{Semantics}: specifies what facts in the world a sentence refers to. It also specifies how you assign a truth value to a sentence based on its meaning in the world. A fact is a claim about the world, and may be true or false.
        \item \textbf{Inference Procedure (reasoning)}: mechanical method for computing (deriving) new (true) sentences from existing sentences.
    \end{itemize}

 
\textbf{Propositional logic (PL)} is a formal system for reasoning with statements that are either true or false, called propositions. It combines these statements using logical connectives such as and ($\land$), or ($\lor$), not ($\neg$), implies ($\implies$), and if-and-only-if ($\iff$). In rule-based systems, PL provides a simple framework for encoding knowledge as facts and if–then rules. Each rule is an implication whose antecedent is a combination of propositional conditions and whose consequent asserts a new proposition or triggers an action. Inference engines apply forward or backward chaining to evaluate conditions and fire applicable rules, often using pattern-matching networks (e.g., Rete \cite{Forgy_1982}) and conflict-resolution strategies to choose which rule executes next. This yields fast, transparent reasoning suitable for business rules and classic expert systems. Its main limitation is expressiveness: without variables or quantifiers, it cannot naturally capture relations or generalizations.

\textbf{First-order logic (FOL)} addresses these limitations by adding variables, predicates, and quantifiers, letting rules express relations and general laws rather than flat, instance-specific facts. A single rule can apply to all matching objects via unification, greatly reducing duplication and aligning well with structured knowledge (e.g., ontologies, relational data). The trade-off is higher computational complexity, so practical systems often use decidable fragments (e.g., Datalog, description logics) or bounded domains. As an example, the following rule expressed in PL:
\begin{multline*}
If \ HighVib^{M_1} \land HighTemp^{M_1} \ then \ Fault^{M_1}; \\
\text{repeat for } M_2, \ M_3, \ ..., \ M_n \\
\text{ where }  M_n  \text{ is  an  individual  machine}.
\end{multline*}
 
\noindent
can be more concisely expressed (generalized) in FOL with the use of a variable $m$.
\begin{multline*}
HighVib(m) \land HighTemp(m) \implies Fault(m), \\
\text{which applies to any machine } m.
\end{multline*}

FOL is by no means the only choice, but as per \cite{Brachman_Levesque_2004} it is a simple and convenient one for the sake of illustration. Real-valued logics (also referred to as ``fuzzy logics'') are often utilized in machine learning because they can be made differentiable and/or probabilistic \cite{Serafini_dAvila_Garcez_2016}. These were first introduced by Łukasiewicz at the turn of the 20th century \cite{McCall_1973,Harder_Besold_2018}. Conjunctions and disjunctions are generalized by the triangular norm (t-norm) and triangular conorm (t-conorm) functions respectively. Table \ref{tab:t-norms} lists several variants of how logical operators can be translated into their real-valued equivalents.

\begin{table}[ht]
\centering
\small
    \begin{tabular}{clll}
        \toprule
        Op. & Gödel & Łukasiewicz & Probabilistic\\
        \midrule
        
        $\land$ & $min\{A,B\}$ & $max(0, \ A + B - 1)$ & $A \cdot B$\\
        $\lor$ & $max\{A,B\}$ & $min(1, \ A + B)$ & $A + B - A \cdot B$\\
        $\neg$ &  $0 \ for \ A > 0$ & $1 - A$ & $0 \ for \ A > 0$\\
        $\implies$  & \begin{math}
                      \Biggl \lbrace 
                      \begin{array}{l@{}l}
                        1 \ if \ A \leq B \\
                        B \ otherwise
                      \end{array}
                      \end{math} 
                      & $min\{1,1-A+B\}$ & 
                      \begin{math}
                      \Biggl \lbrace 
                      \begin{array}{l@{}l}
                        1 \ if \ A \leq B \\
                        B/A \ otherwise
                      \end{array}
                      \end{math}\\
        \bottomrule
    \end{tabular}
\caption{Logical operators and their real-valued logic expressions \cite{Metcalfe_2005}.}
 \label{tab:t-norms}
\end{table}

Choice of t-norms and t-conorms depends on the application. Idempotent\endnote{A mathematical quantity which when applied to itself under a given binary operation (such as multiplication) equals itself.
} operators—minimum and maximum—are appropriate when retention of original values is required. Strict operators—e.g., the product t-norm and the probabilistic-sum t-conorm—yield outputs that are lower (for t-norms) or higher (for t-conorms) than the inputs for interior values, supporting smoother, more compensatory aggregation. Nilpotent\endnote{An algebraic quantity that when raised to a certain power equals zero.} operators—such as the Łukasiewicz pair—can return 0 (t-norm) or 1 (t-conorm) for certain inputs, which is useful when modeling complete non-membership or full membership. 

While a comprehensive exposition of real valued logics is beyond the scope of this review, we refer the interested reader to \cite{Shakarian_2023}, which is of particular relevance to Neuro-symbolic AI.


  
\textbf{Temporal logic (TL)} is a family of formal logics for describing and reasoning about how the truth of propositions changes over time. It was originally conceptualized by Arthur Prior in the 1950s under the moniker of ``tense logic'', while Jerzy Łoś constructed, described and examined the first mature TL calculus \cite{sep-prior,Parol_2021}. TL extends PL or FOL with temporal operators that express constraints such as:

\begin{itemize}
    \item \textit{G (always/globally)} - a property holds at all times.
    \item \textit{F (eventually)} - a property will hold at some future time. 
    \item \textit{X (next)} - a property holds at the next time step.
    \item \textit{U (until)} - one property holds until another becomes true.
\end{itemize}

\noindent
often with optional time bounds. Some of the main variants of TL include: Linear-Time Temporal Logic (LTL) for dealing with discrete sequences of states; Computation Tree Logic (CTL/CTL*) for branching futures; Interval/Action logics (ITL/TLA+) for intervals and state transitions; and Metric Temporal Logic (MTL) for explicit time bounds on the operators eventually/always/until over discrete time. 

Especially applicable to time-series data, \textbf{Signal Temporal Logic (STL)}) \cite{Maler_Nickovic_2004} is a formal language for specifying and monitoring properties of real-valued, time-varying signals, common in cyber-physical systems. It extends temporal logic with time-bounded operators (always, eventually, until) and predicates over signals (e.g., temp $<$ 80), allowing requirements like \textit{temperature stays below 60 degrees for the first 10 seconds} or \textit{whenever vibration spikes, pressure drops within 2 seconds}. STL supports quantitative ``robustness'' semantics that measure how strongly a signal satisfies or violates a spec, enabling optimization-based testing and control. It is widely used for runtime monitoring, falsification, and controller/parameter synthesis.

This formalism is particularly promising in the PdM domain where complex expert defined rules can be represented using logical formulas. Furthermore, these formulas can be expressed using differentiable functions which can be used in the design of a neural network architecture. For example, the rule: \textit{Zone temperature is lower than setpoint by 2 degrees Celcius or more when heating valve is fully open ($>98\%$) and cooling valve is fully closed ($<2\%$) and Fan status is On in occupied mode. These conditions were true for more than 60 minutes (2 or more times in a day), or more than 120 minutes once in a day}, can be represented using STL formulas and used as prior knowledge for training a fault detection model. Several works have been proposed in this direction \cite{Raman_2017,NSTSC_2022,Tian_2024,Stl2vec_2024, GradSTL_2025,STLCG++_2025}.


For completeness, other, logic-based cognitive modeling approaches such as non-monotonic logic, attempt to deal with the complexities of human reasoning, epistemology, and defeasible inference \cite{sep_logic_nonmonotonic}. Natural Logic (NL) is a formal proof theory built on the syntax of human language, which can be traced to the syllogisms of Aristotle \cite{Byszuk_Wozniak_2020}.

\subsubsection{Ontologies and Knowledge Graphs (KG)} - In predictive maintenance, an ontology is a formal, explicit, machine-interpretable specification of a shared conceptualization of equipment health and maintenance. It defines the domain’s core classes (e.g., assets, components, sensors, observations, operating conditions, units, failure modes, health indicators, prognostics such as remaining useful life, maintenance tasks, resources) and their relationships, constraints, and rules. This canonical vocabulary enables semantic integration of heterogeneous Operational and Information Technology OT/IT data, supports reasoning and analytics for fault detection, diagnosis, and decision support, and facilitates interoperable exchange across monitoring, control, CMMS, Enterprise Asset Management (EAM), and digital-twin systems.

Ontologies separate terminological knowledge (T-box: taxonomy/vocabulary), or schema, from assertional knowledge (A-box: facts about instances), or instances, commonly represented in Web Ontology Language (OWL) to ensure logical consistency and interoperability. In practice, an ontology is often referred to as the schema, while a Knowledge Graph (KG) comprises the data instances. However, a KG can exist on its own without an explicit ontology having been defined a-priori. Such KGs are sometimes referred to as property graphs which contain domain knowledge, but which cannot be reasoned over. In the literature the terms ``ontology'' and ``knowledge graph'' are sometimes used interchangeably.

\begin{figure}[ht]
    \centering
    \includegraphics[width=1\linewidth]{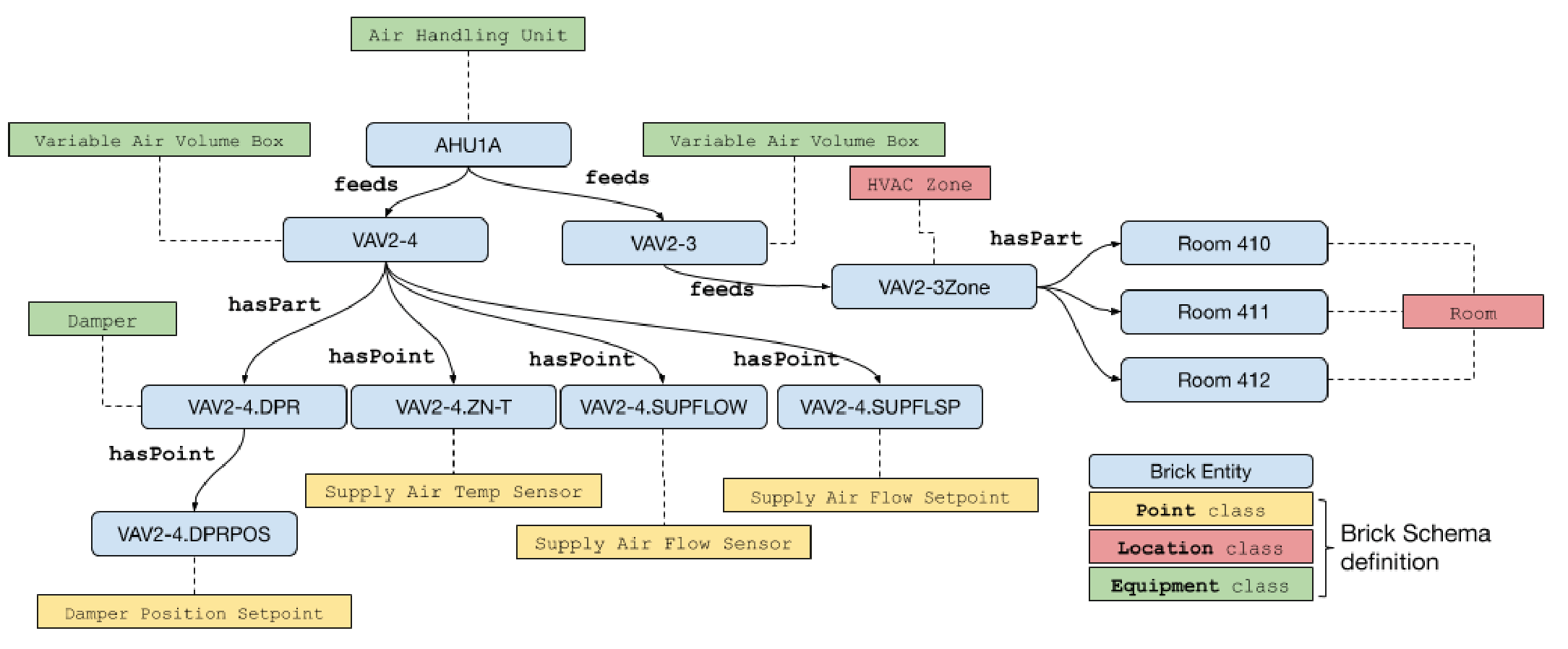}
    \caption{Example of a Brick model reproduced from \url{brickschema.org}. A Brick model is a digital representation of a building that adheres to the Brick schema.}
    \label{fig:brick}
\end{figure}

To improve maintenance operations, ontologies frequently use Semantic Web Rule Language (SWRL) rules and case-based reasoning systems to analyze sensor data and historical records \cite{Fernandez_2024}. For instance, if there is a failure in an HVAC system, the ontology should allow for tracing back the possible causes, which could range from environmental factors to previous maintenance activities. One of the guiding principles of ontology development is the notion of reuse. Thus ideally, domain ontologies should implement standards and be made publicly available. One such ontology in the maintenance domain is the \textit{Brick} schema\endnote{\url{https://brickschema.org/}}. An illustrative example of \textit{Brick} is given in Figure \ref{fig:brick}. For a comparison of similar technologies we refer the reader to the \textit{Brick} website. A great deal of expertise is required in both the domain and ontology design. \cite{Gispert_2025} warn that if relationships are not well defined, it becomes difficult to perform root cause analysis and make informed predictions. This difficulty may contribute to the low rate of adoption. Of the 3,000+ studies selected for this review, only 18 mention either ontologies or KGs.


\subsubsection{Baysian Networks}\label{bayesian-networks}
For root cause analysis, ontologies can be combined with Causal Bayesian Networks (CBN) \cite{Jaimini_Henson_Sheth_2023}. CBNs encode cause–effect structure among well-defined variables and support interventional and counterfactual reasoning using do-calculus \cite{Pearl_2012}.\endnote{Do-calculus was developed in 1995 by Judea Pearl to facilitate the identification of causal effects in non-parametric models \cite{Pearl_2012}}. Nodes represent factors like loads, process conditions, health indicators, and failure modes; edges capture causal influence (e.g., low NPSH\endnote{NPSH stands for Net Positive Suction Head, which is a measure of the pressure available at the suction side of a pump to prevent cavitation.} causes cavitation; cavitation accelerates seal wear). Used together, ontologies ground a CBN in precise semantics and high-quality variables. Practically, the CBN graph and parameters can be represented in a knowledge graph (using RDF/OWL languages) as instances (nodes, edges, CPDs\endnote{Conditional Probability Distribution (CPD) is used in Bayesian Networks to define the probabilities associated with each node based on its parent nodes.}) linked to the ontology’s classes, while probabilistic inference runs in a BN engine.

\cite{Taal_Itard_2020} propose The Diagnostic Bayesian Network (DBN) for Automated Fault Isolation (AFI) that uses Bayes theory to estimate fault probabilities based on detected symptoms. The nodes are categorized into parent fault nodes (representing the three types of faults such as model, component, and control) and child symptom nodes (representing the four symptom types such as balance, energy performance (EP), operational state (OS), and additional symptoms). The edges (arrows) are directed from the faults (causes) to the symptoms (effects), with relationships established based on HVAC Process and Instrumentation Diagrams (P\&IDs). Modeling parameters include prior probabilities for fault nodes and conditional probabilities for symptom nodes (often using Noisy-MAX nodes), which are set once-off, based on HVAC expert knowledge. Inference is performed by inputting observed symptom data to calculate the posterior fault probabilities. The authors report a 25\% reduction in energy consumption.

There are 13 studies that explicitly mention BNs. Another 63 mention other Bayesian techniques. Of these, two studies \cite{Hosamo_Kofoed_2023a,Hosamo_Kofoed_2023b} combine the BNs with KGs to assist decision making through visualization. As an aside, another 63 mention Bayesian techniques and inference but the techniques are not specified.  Table \ref{tab:ontology-cbn} highlights several examples of the combination of ontologies and BNs and how they can be used.

In general, graphical structures offer a promising way forward, as they naturally represent heterogeneous data and system topology while supporting structured inference (including logical reasoning). Industrial systems typically consist of many interdependent elements—such as motors, pumps, gearboxes, valves, shafts, bearings, seals, and control units — yet much of the PdM literature still treats each element (e.g., a single rolling bearing) in isolation. Graph-based representations are especially well-suited to neuro-symbolic approaches, since symbolic expressions (such as rules, causal relations, or temporal logic formulas) can be encoded as graph structures and integrated with learned embeddings \cite{lamb_garcez_gnn_2021}. For example, combining CBNs with ontologies for root cause analysis (RCA) may offer a promising route to addressing a difficult yet very high-impact research area. Admittedly, a major challenge lies in constructing a domain-specific ontology that is both general enough to be reusable and detailed enough to capture the nuances of a specific application. Automatic ontology construction is a research field in its own right, and we believe the PdM community could greatly benefit from closer collaboration with experts in this area.


\subsubsection{Case-based reasoning (CBR)}\label{sec:cbr} solves new maintenance problems by retrieving and adapting solutions from similar past situations, or cases. In predictive maintenance, a ``case'' typically bundles the asset and configuration, operating context (speed, load, environment), symptoms and sensor fingerprints (e.g., vibration bands, temperature trends, oil debris), the diagnosed failure mode, actions taken, parts used, time-to-failure or RUL, and the eventual outcome. When a new alert is triggered, the system computes similarity to past cases using engineered features, domain ontologies, or learned embeddings to surface the most analogous scenarios and their successful remedies. This yields an explainable recommendation such as ``outer race defect likely; replacement within 5 days minimized downtime in 8/10 similar cases,'' along with expected effectiveness, lead times, and risks.

CBR is especially valuable when labeled failure data are sparse but historical work orders and Subject Matter Expert (SME) knowledge exist. CBR pairs well with knowledge graphs (for context and retrieval) and probabilistic models (for calibrated risk), and is evaluated by recommendation accuracy, time-to-diagnosis, avoided downtime, and cost savings. \cite{Klein_2025} show that if domain knowledge is perfectly modeled, working directly at the symbolic level is advantageous. However, if there are gaps in the modeled knowledge, the approaches using embeddings and Neuro-symbolic approaches in which the weaknesses of purely symbolic solutions are mitigated while retaining the strengths.

For time-series data, Temporal Case-Based Reasoning (TCBR) \cite{Jare_2002} extends CBR by representing cases as time-dependent episodes—sequences of states, sensor trends, and events with their timing and durations—rather than static snapshots. It retrieves and aligns similar timelines to the current evolving trajectory to forecast what will happen next and recommend timely actions. Apart from \cite{Schultheis_2024}'s research proposal, very few studies exist which utilize TCBR in PdM. The following are ones we were able to find.

\cite{May_Cho_2022} presented an ontology-enabled case-based reasoning framework to enhance predictive maintenance architectures. By combining semantic domain ontologies with CBR, their approach improves the retrieval and reuse of past failure cases for diagnosing equipment anomalies and planning maintenance actions in advance.

\cite{Colloc_2022} developed a temporal case-based reasoning (TCBR) platform that extends an object-oriented fuzzy vector space model for comparing cases based on their evolution over time. Their method computes distances between temporal fuzzy vector representations of case attributes to measure similarity in how a system’s state changes over time, in order to identify similar degradation patterns.

\cite{Rodríguez_2024} integrated CBR with the Reliability-Centered Maintenance (RCM) methodology \cite{Nowlan_Heap_1978}. This combination of RCM and CBR showed improved efficiency (e.g. time savings in failure mode analysis) by reusing past cases to anticipate and prevent equipment failures.

\subsubsection{Fault trees (FTs)} 
\begin{figure}[ht]
    \centering
    \includegraphics[width=1\linewidth]{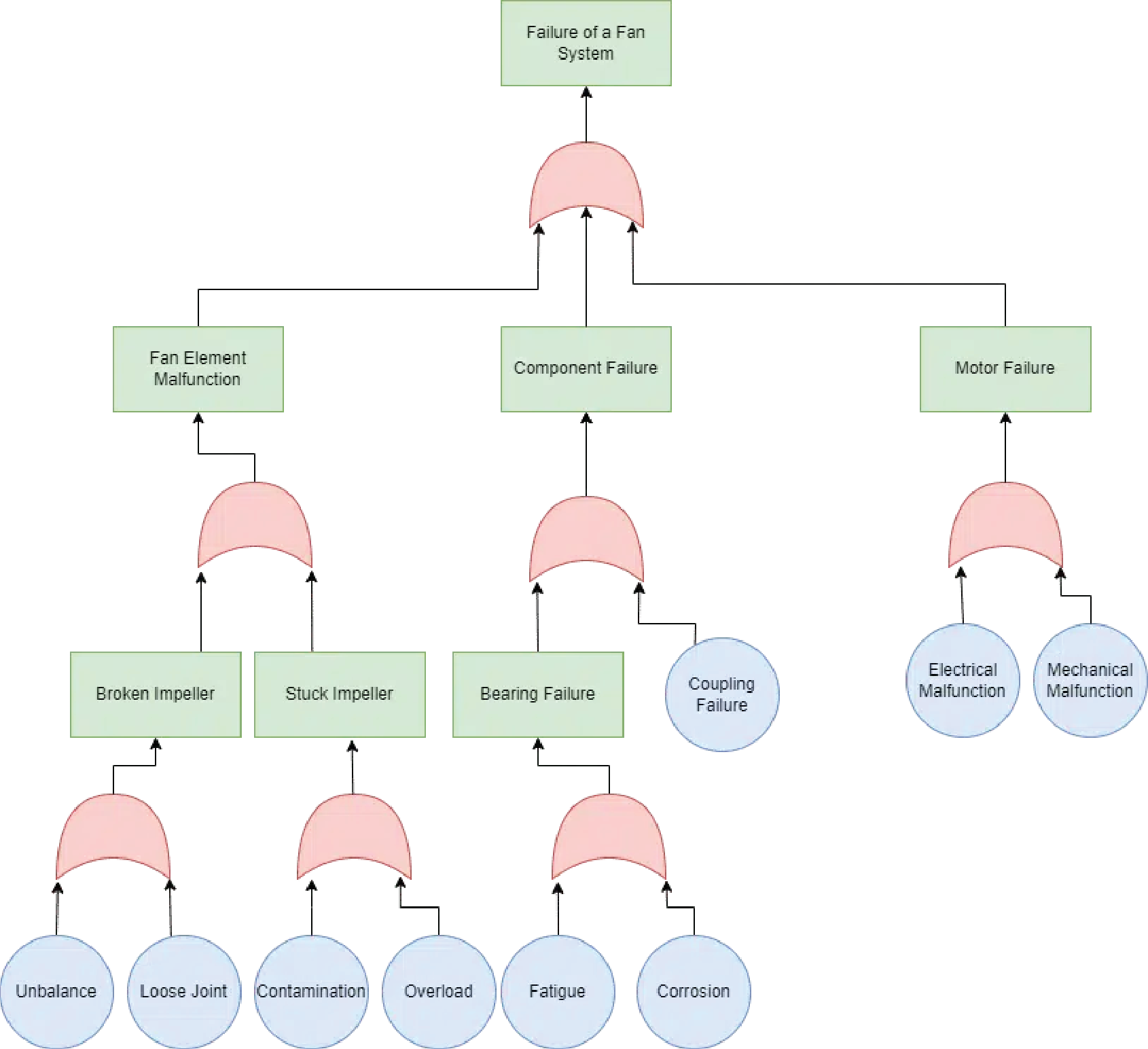}
    \caption{Fan System Failure Diagram reproduced from \url{https://sensemore.io/what-is-fault-tree-analysis/.}}
    \label{fig:fault-tree}
\end{figure}

(used in fault tree analysis, or FTA) are another type of graphical model that maps out how various sub-events or component failures can combine to cause a larger system failure. In a fault tree diagram, the undesired top event (e.g. a machine breakdown) is at the root, and contributing faults or conditions branch below, connected by logical gates (such as AND and OR gates - classical boolean logic) indicating how combinations of those events lead to the top failure. Fault trees help visualize and quantify risk. \cite{Xie_Lu_2020} argue that a visual tree representation of faults helps both users and professionals in filtering and finding the main cause of anomalies, and that the visualized inspection system could also make it easier for site workers to locate and repair failed assets when conducting maintenance tasks. Figure \ref{fig:fault-tree} shows an example of a fault-tree diagram of a fan system failure. There are 5 studies utilizing FTs \cite{Aghajanian_Rao_2021,Mandelli_2024,Niloofar_2023,Rahman_2023,Xie_Lu_2020}.

Knowledge-based systems are transparent to users, and good at capturing operational context. However, they tend to be brittle outside the encoded scenarios, costly to maintain as equipment and processes evolve, and weak at predicting novel or physics-driven failure modes. To address the limitations of physics-based and knowledge-based techniques, data-driven approaches have gained popularity.


\subsection{Data-driven} \label{sec:data-driven}

\begin{figure}
    \centering
    \includegraphics[width=1\linewidth]{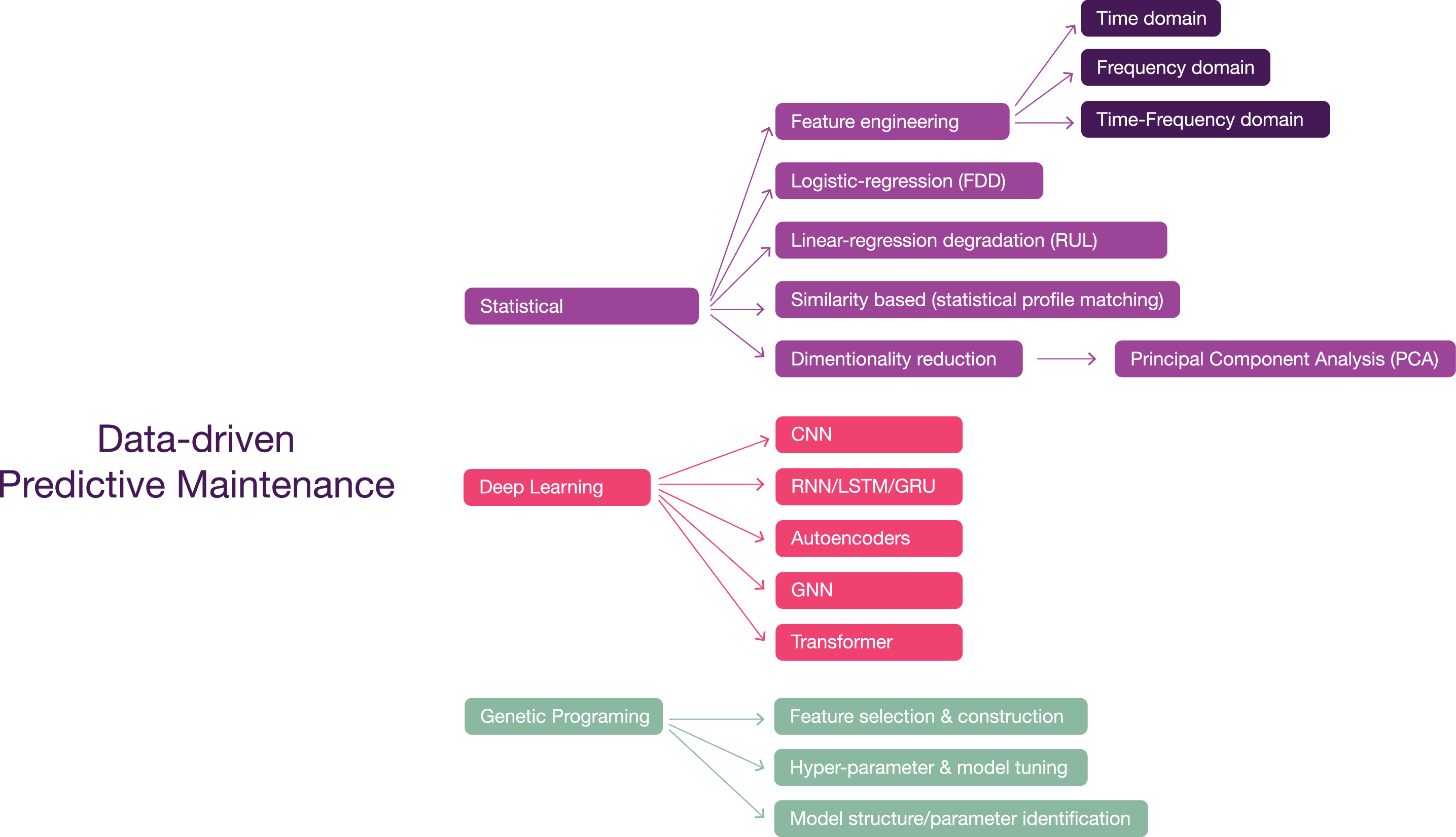}
    \caption{Types of Data-driven Predictive Maintenance (PdM).}
    \label{fig:data-driven-diagram}
\end{figure}

The majority of the PdM literature reviewed in this study utilizes one or more data-driven modeling techniques. These can be broadly categorized into statistical, genetic algorithms and programming, and machine learning/deep learning. This section introduces each category with examples where applicable - Figure \ref{fig:data-driven-diagram}.

\phantom{TODO: find the correct subsection for this citation. \cite{Irani_2023} train a model using the normal operational data from an HVAC system. The realized model is used to design the parity space and generate the residuals for diagnosing sensor malfunctions in the system. The proposed method does not require any prior knowledge regarding the HVAC dynamics and only uses the data collected from the normal operation of the system. }

\subsubsection{Statistical Signal Processing (SSP)}
Before the rise of modern Machine Learning (ML) techniques, statistical analysis was the primary method used in the manufacturing industry \cite{Sheuly_Barua_2021}. As the name suggests, SSP uses statistical characteristics for converting raw, complex, and noisy sensor data into meaningful parameters. SSP is integral to several stages of data analysis, from generating  features, to developing statistical models for monitoring system degradation and uncertainty, to dimensionality reduction. Some advantages of statistical methods include: 

\begin{itemize}
    \item \textbf{Generalizability}: Statistical methods require no mechanistic understanding of the underlying physical process, making them generalizable across different equipment.
    \item \textbf{Data scarcity}: Statistical methods require limited historic data compared to other methods.
    \item \textbf{Data augmentation}: Degradation models quantify how a machine’s condition changes over time and are used for RUL prediction. Synthetic degradation data can be generated using models that incorporate stochastic variables to mimic real-world variance. 
\end{itemize}

The literature highlights three distinct statistical model types used in PdM: linear-regression degradation models, similarity-based models that match against prior cases, and logistic-regression models estimating failure probability. All three rely on statistical features that serve as the models’ parameters.

\textbf{Degradation statistical models} forecast Remaining Useful Life (RUL) by performing regression on a machine's Health Indicator (HI) data to predict when degradation will reach a specified failure threshold \cite{Harries_Hartnoll_2023}. A common technique is the exponential degradation model \cite{Li_Zhao_Fu_Cao_2024}, which is refined through iterative updates based on new data from subsequent asset use cycles (cutting cycles in the case of cutting tools) and feedback from previous prediction errors. The model dynamically updates parameters to calculate the probability density distribution and create a rolling RUL mapping. In a digital twin of a bicycle factory, \cite{Harries_Hartnoll_2023} show superior performance of the degradation model over Time-Based Maintenance (TBM), yielding a mean improvement of up to 12.9\% for linear degradation profiles. However, a major limitation is that the model’s effectiveness is reduced by increased noise in the signal.

\textbf{Similarity statistical models} estimate RUL by comparing the current machine performance data to a historic data ensemble of previous failure profiles. This is conceptually similar to case-based reasoning discussed above. The core technique involves seeking which historical profiles most closely match the current machine’s profile and then calculating the RUL by averaging the predicted remaining lives of the closest matches. An advantage is that this model often proves more successful across a wider range of parameter permutations than the degradation model  \cite{Harries_Hartnoll_2023}. Furthermore, \cite{Harries_Hartnoll_2023} show that its performance for linear wear profiles improves with reduced noise and a smaller historical ensemble size. However, this model’s limitation is its strong reliance on the historical dataset; consequently, if the new operational profile differs significantly from historical failure examples, the prediction will be less effective.

\textbf{Statistical Models for Failure Probability} are based on multivariate logistic regression trained on selected sensor features (such as current mean or RMS) to calculate a failure conformance value (CV), or equipment health index, mapped between 0 (healthy) and 1 (faulty). \cite{Negri_2021} utilize this technique in an Equipment Prognostics and Health Management (EPHM) module within a Digital Twin framework to quantify equipment health and uncertainty in real-time for production scheduling. The primary advantage is that this failure CV serves as the quantified failure probability input for subsequent Discrete Event Simulation (DES) models. A limitation of this approach is that the supervised learning method used assumes the signature of the failure in the collected signals is already known.

With the rise of Machine Learning (ML), statistical methods are used mainly to derive signal features and to aid dimensionality reduction to be used as input to ML models, rather than model design itself. Extracting appropriate features from raw signals is necessary because they often exhibit characteristics like large data volume, high dimensionality, and noise. In the context of feature extraction, statistical methods provide quantifiable metrics that reflect characteristic values of signal variation in the time domain, the frequency domain, and the time-frequency domain. \cite{Zhang_Wang_2025} provide a thorough analysis of these signal characteristics of control sensors within HVAC systems.

\begin{itemize}
    \item \textit{Time-Domain} features are derived using statistical methods to indicate signal amplitude and energy, such as mean, RMS, and standard deviation. Higher-Order Statistics and Distribution Metrics are used to analyze the shape and distribution of the sensor signals such as kurtosis and skewness. Time-domain features and their formulas are given in Table \ref{tab:time-dom-features}.
    \item \textit{Frequency-Domain} features are used to effectively capture periodic components and the spectral characteristics of the sensor data. These are generally derived from signal processing techniques like the Fast Fourier Transform (FFT) and Power Spectral Density (PSD) analysis. The useful information in sensor signals is primarily concentrated in the low-frequency range, while noise is mainly distributed in the high-frequency range \cite{Zhang_Wang_2025}. Frequency-domain features and their formulas are given in Table \ref{tab:freq-dom-features}.
    \item \textit{Time-Frequency-Domain} features are analyzed by incorporating techniques like Wavelet Transform (WT) and Wavelet Packet Decomposition (WPD). This method is particularly useful because it can adaptively decompose a very noisy signal into a collection of intrinsic mode functions (IMFs) \cite{Wang_Yang_Guo_2022}. IMFs represent simple oscillatory modes and allow for the extraction of instantaneous frequency and amplitude information, making them useful in vibration analysis to detect faults in machinery. IMFs are generated through a process called empirical mode decomposition (EMD), which breaks down a signal into its constituent parts without losing time-domain information. The choice of IMF is evaluated using Statistical Complexity Measures (SCM). The formula for the time-frequency-domain feature Intrinsic Mode Energy is given in Table \ref{tab:time-freq-dom-features}.
    

\end{itemize}

Finally, for dimensionality reduction, signal processing commonly employs Principal Component Analysis (PCA) and its variants. PCA is typically implemented using Singular Value Decomposition (SVD) to extract the principal components (eigenvectors) in the data. The first component explains the largest share of variance, the second the next largest, and so on. Consequently, the top n components capture most of the variance. Projecting data onto the top few components reduces dimensionality while preserving information, yielding uncorrelated features and often improving visualization, compression, and noise reduction.

PCA methods extended to handle uncertain, imprecise data (known as Interval PCA or IPCA) are popular, including Midpoint-Radius PCA (MRPCA) and Complete Information PCA (CIPCA). However, traditional IPCA methods often only capture the global information (data variance), neglecting local neighborhood information. To explicitly addresses this limitation \cite{Li_Ding_Sun_Liu_Pu_2024} propose Local and Global Interval Embedding Algorithms (LGIEA)  by combining CIPCA (which maximizes global scattering/variance) and ILPP (which minimizes local scattering), thereby extracting both global and local feature information from interval-valued data. The proposed method significantly reduces the error alarm rate and the false monitoring rate and improves the accuracy of data classification compared to traditional PCA methods.

\begin{table}[ht]
    \centering
    \begin{tabular*}{\linewidth}{@{\extracolsep{\fill}}ll}
    \toprule
        Feature  & Formula  \\
    \midrule    
        Mean  & 
        $
        T_{avg} = \mu_t = \frac{1}{n} \sum_{i=1}^{n} s_i
        $ \\
        Root Mean Square &
        $
        T_{rms} = \sqrt{\frac{1}{n} \sum_{i=1}^{n} s_i^2}
        $ \\
        Standard Deviation &
        $
        T_{std} = \sigma_t = \sqrt{\frac{1}{n - 1} \sum_{i=1}^{n} (s_i - \mu_t)^2}
        $ \\
        Shape factor &
        $
        T_{shape} = \frac{T_{rms}}{\frac{1}{n} \sum_{i=1}^{n}}
        $ \\
        Kurtosis &
        $
        T_{kurt} = \frac{1}{\sigma_t^4} \frac{1}{n} \sum_{i=1}^{n} (s_i - \mu_t)^4
        $\\
        Skewness &
        $
        T_{sk} = \frac{1}{\sigma_t^3} \frac{1}{n} \sum_{i=1}^{n} (s_i - \mu_t)^3
        $ \\
        Crest factor &
        $
        T_{crest} = \frac{T_{peak}}{T_{rms}}
        $ \\
        Impulse factor &
        $
        T_{impulse} = \frac{\max(s)}{\frac{1}{n} \sum_{i=1}^{n}}
        $ \\
        Clearance factor &
        $
        T_{clear} = \frac{\max(s)}{\frac{1}{n} \sum_{i=1}^{n} }
        $ \\
        Peak value &
        $
        T_{peak} = \max(s)
        $ \\
        Signal-to-Noise and Dist. Ratio &
        $
        T_{SINAD} = 10\log_{10} \frac{P_{signal} + P_{noise} + P_{distortion}}{P_{noise} + P_{distortion}}
        $ \\
        Total Harmonic Distortion &
        $
        T_{THD} = 10\log_{10} \frac{\sum_{n=2}^{N} P_n}{P_1}
        $ \\
        Signal-to-Noise Ratio &
        $
        T_{SNR} = 10\log_{10} \frac{P_{signal}}{P_{noise}}
        $ \\
    \bottomrule
    \end{tabular*}
    \caption{Time-domain features. \\
    $s=[s_1,s_2,\ldots,s_n]$ represents the measured signal, and n is the signal length. \\
    $\mu_t$ is the mean value of the time-domain signal, and $\sigma_t$ is the standard deviation. \\
    $P$ terms $(P_{signal},P_{noise},P_{distortion})$ refer to the power of the respective components within the signal. $P_1$ is the power of the fundamental harmonic, and $\sum_{n=2}^N P_n$ is the sum of the power of the second to $N^{th}$ harmonics.}
    \label{tab:time-dom-features}
\end{table}

\begin{table}[ht]
    \centering
    \begin{tabular*}{\linewidth}{@{\extracolsep{\fill}}ll}
    \toprule
        Feature & Formula \\
    \midrule
        Mean frequency & 
        $
        f_{avg} = \mu_f = \frac{1}{n} \sum_{i=1}^{n} S(\omega_i)
        $\\
        Frequency standard deviation & 
        $
        f_{std} = \sigma_f = \sqrt{\frac{1}{n - 1} \sum_{i=1}^{n} ( S(\omega_i) - \mu_f )^2}
        $\\
        Frequency skewness & 
        $
        f_{sk} = \frac{1}{n} \sum_{i=1}^{n} \frac{( S(\omega_i) - \mu_f )^3}{\sigma_f^3}
        $\\
        Peak amplitude & 
        $
        f_{peakA} = \max(|S(\omega)|)
        $\\ 
        Peak frequency & 
        $
        f_{peakf} = f_{Wn} = S^{- 1} ( f_{peakA} )
        $\\
        Zeta & 
        $
        f_{Zeta} = \frac{f_2 - f_1}{2 f_{Wn}}
        $\\
    \bottomrule
    \end{tabular*}
    \caption{Frequency-domain features. The frequency-domain features $(f)$ are derived from signal processing techniques like the Fast Fourier Transform (FFT) and Power Spectral Density (PSD) analysis. These features are used to capture periodic components and the spectral characteristics of the sensor data.}
    \label{tab:freq-dom-features}
\end{table}

\begin{table}[ht]
    \centering
    \begin{tabular*}{\linewidth}{@{\extracolsep{\fill}}ll}
    \toprule
    Feature & Formula \\
    \midrule
         Intrinsic Mode Energy &  
         $
         E_{i,j} = \sum_{k=1}^{n} \left| d_{i,j}(k) \right|^2, n = 0, 1, \ldots
         $\\
    \bottomrule
    \end{tabular*}
    \caption{The Time–Frequency Domain analysis incorporates techniques like Wavelet Transform (WT) and Wavelet Packet Decomposition (WPD). $d_{i,j}(k)$ represents the percentage of each node derived from the wavelet packet decomposition. The terms $i$ and $j$ denote the node number and decomposition layer, respectively.}
    \label{tab:time-freq-dom-features}
\end{table}

\subsubsection{Genetic algorithms (GAs)} 
\begin{figure}[ht]
    \centering
    \includegraphics[width=0.75\linewidth]{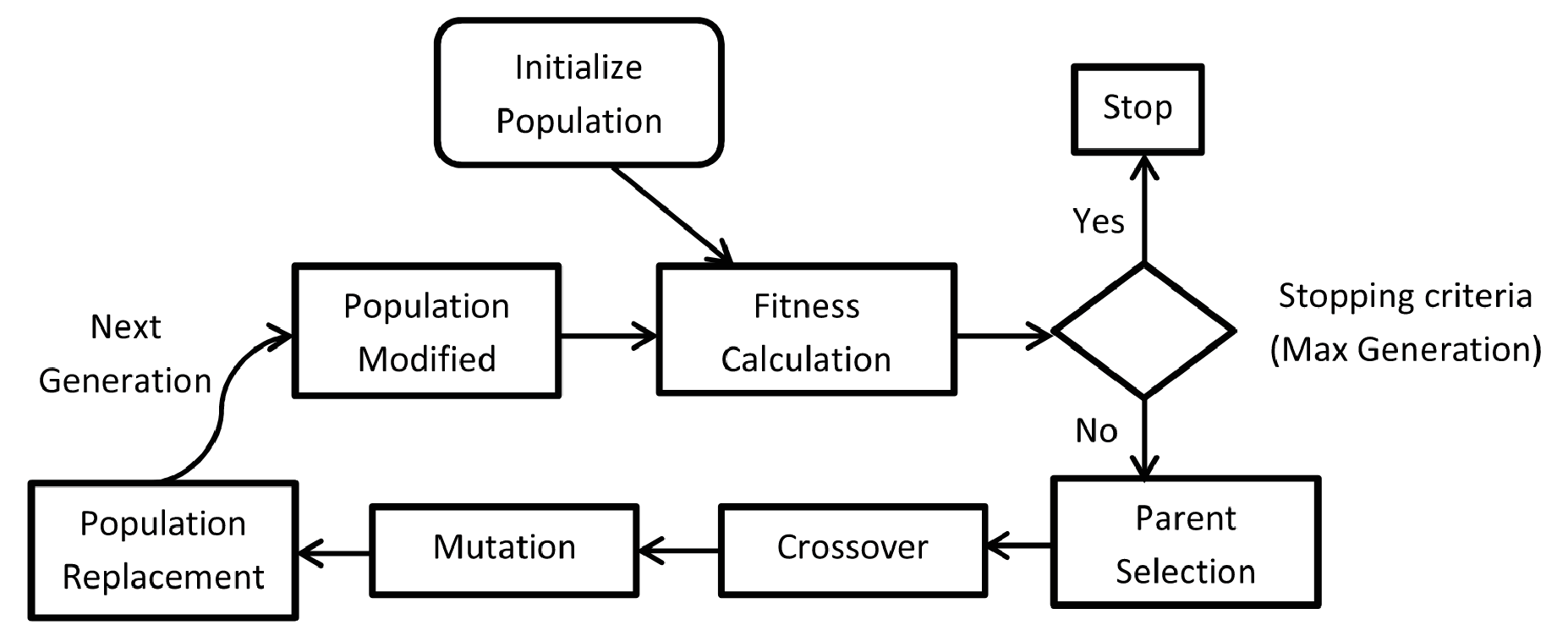}
    \caption{Steps in Genetic Algorithms, reproduced from \cite{Kumar_Mukherjee_2022}}
    \label{fig:genetic-algo}
\end{figure}

are population-based, stochastic optimization methods inspired by natural selection. They evolve a set of candidate solutions (chromosomes) by repeatedly evaluating a fitness function, selecting fitter candidates, recombining them (crossover), and introducing random changes (mutation), often with ``elitism'' to retain the best. Over generations, the population converges toward high-quality solutions, including for problems that are nonconvex, discrete, or have complex constraints. Figure \ref{fig:genetic-algo} depicts the steps involved in GAs. GAs can be used for feature selection and construction, hyperparameter and model tuning, or model structure/parameter identification. A brief description of each follows.

To perform \textbf{Feature selection and construction}, the most informative features are chosen from time-series data, by evolving wavelet scales, filter bands, or statistical descriptors to maximize diagnostic accuracy while minimizing feature count.

\cite{Jaen-Cuellar_2023} use GAs to optimize the selection of statistical features for detecting gradual gear wear. The core strategy here is based on maximizing the variance of the features (VF), which serves as the fitness value. High variance indicates that the features have high dispersion and can be better separated for fault conditions. This approach improved overall diagnostic performance to 92.2\% when fused with Linear Discriminant Analysis (LDA), significantly surpassing results from analyses without feature optimization (e.g., 72.5\% for current signals alone). Interestingly, when GA optimization was applied across multiple cases, the most commonly selected features were the mean, Root Mean Square (RMS), and high-order moments, which can have direct physical meaning. 
    
\cite{Lee_Le_2020} introduce Genetic-Binary State Transition Algorithm (GBSTA) for feature selection. GBSTA combines the strong global search (exploration) of GA (using the crossover operator) with the local exploitation capabilities of the Binary State Transition Algorithm (BSTA) to achieve a balance of exploitation and exploration in the search space, which is often difficult for standalone algorithms. In BLDC\endnote{A brushless DC motor (also known as a BLDC motor or BL motor) is an electronically commuted DC motor which does not have brushes. The controller provides pulses of current to the motor windings which control the speed and torque of the synchronous motor.\url{https://www.electrical4u.com/brushless-dc-motors/}} motor bearing fault classification, GBSTA achieved an average accuracy of 99.6\% when combined with an Artificial Neural Network (ANN) classifier.

\cite{Kumar_Anand_2024} explore different feature selection algorithms to train an SVM to classify defects in rolling ball bearings. The overall accuracy of the SVM model is improved to 97\% under no load operation and 95\% under full load operation with the Fisher score algorithm. It is further improved to 99\% under no load operation and 95\% under full load operation showcasing the superior efficacy of GA for feature selection.
    
\textbf{Hyperparameter and model tuning} consists of optimizing model (e.g., SVM, CNN, LSTM) parameters, ensemble weights, threshold levels, or window lengths for anomaly detection and RUL prediction (e.g., tuning LSTM layers and learning rates or selecting SVM kernel/C). 

\cite{Akpudo_Hur_2021} propose Multi-Objective Genetic Algorithm-optimized Long Short-Term Memory (MOGA-LSTM) for solenoid pump RUL prediction. MOGA-LSTM aims to discover the best LSTM parameters that return both the minimum Root Mean Square Error (RMSE) of the RUL prediction and the maximum Relative Accuracy (RA). It is superior to other multi-objective optimizers such as Multi-Objective Particle Swarm Optimization (MOPSO), and shows reliable efficiencies for global parameterization of time-series predictors.
    
\textbf{Model structure/parameter identification} is used to fit degradation or reliability models (Weibull/PHM/covariate models), calibrate physics-informed parameters (e.g., Arrhenius aging constants, friction/leakage terms), or tune estimator covariances (Q/R of Kalman filters) to best match historical data. In particular, symbolic regression (SR) \cite{Schmidt_Lipson_2009} is where a mathematical model is learned from the data by optimizing for model parsimony while minimizing errors. This is done by searching the space of mathematical expressions that best fit the data by combining operators, functions, or constants (i.e. mathematical symbols) through the use of evolutionary algorithms such as Genetic programming (GP). 
    
\cite{Hale_Safikou_Bollas_2022} propose a GP procedure for SR to automatically generate the mathematical functions that define inferential sensors (or soft sensors). These uniquely explainable mathematical functions minimize an objective function (such as the mean squared error). The resulting inferential sensor is an equation that is transparent and interpretable. The presented inferential sensor method was compared to traditional fault classification methods such as the support vector machine and pattern recognition neural network algorithms, where it was found to have equivalent accuracy and robustness to model error.

Most recently however, MetaSymNet \cite{Li_Li_Yu_2025}, treats SR as a numerical optimization problem rather than a combinatorial optimization problem, outperforming other SOTA SR algorithms including DSO \cite{Mundhenk_2021}, TSPR \cite{shojaee2023}, SPL \cite{sun2023symbolic}, NESYmRes \cite{biggio21a}, EQL \cite{MartiusL16}.

In summary, the most promising GA strategies center on selecting the most impactful features (feature selection) or tuning model parameters (hyperparameter optimization) to achieve high accuracy and robustness, often using techniques like maximizing feature variance or combining GA with other optimization/ML methods. It is important to note that the majority of GA studies focus on Feature Selection (FS) or dimensionality reduction, rather than identifying the mathematical model structure or internal prediction parameters. There are 37 studies utilizing GAs, in combination with other data-driven techniques. 



\subsubsection{Deep Learning (DL)} is a subset of Machine Learning (ML) that uses multi-layered neural networks to automatically learn complex representations from raw data (e.g., images, audio, text, or signals), often with minimal manual feature engineering. More broadly, ML includes many other methods—such as decision trees, random forests, and linear models—that typically rely more on hand-crafted features and shallower architectures, and may be easier to interpret but less effective on very large, high-dimensional datasets. Given deep learning’s ability to extract patterns from large volumes of data with high accuracy, it is unsurprising that most recent PdM studies employ some form of deep learning algorithm.

\begin{figure}[ht]
    \centering
    \includegraphics[width=1\textwidth]{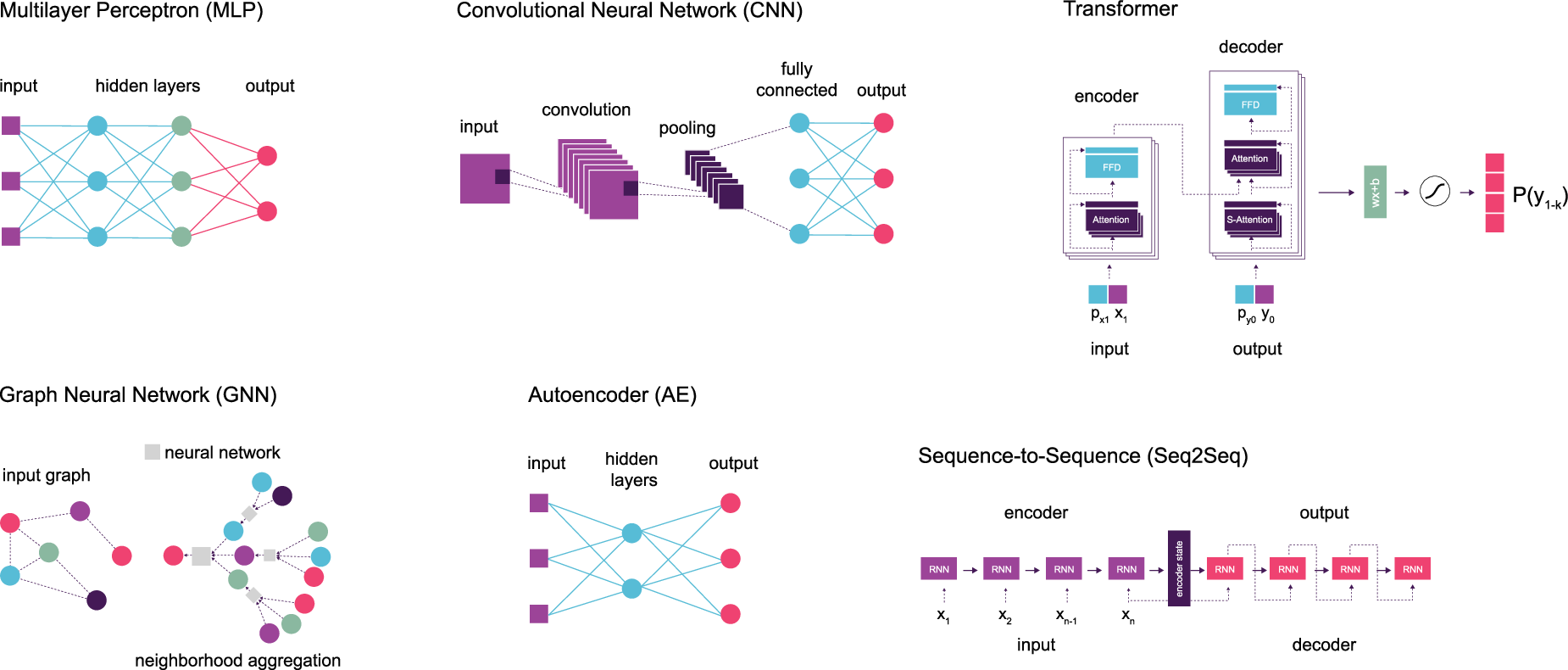}
    \caption{Major categories of neural architectures.}
    \label{fig:neural-arch}
\end{figure}

Deep Learning can be categorized into 1) the type of learning: supervised, unsupervised, and reinforcement learning, 2) the type of output: discriminative and generative, and 3) the type of architecture, starting from the most simple Feed Forward Neural Network (FFNN) to the most recent advancements in Transformer based models. Six common categories of neural architectures are illustrated in Figure \ref{fig:neural-arch}: Multilayer Perceptron (MLP) \cite{Rosenblatt1958ThePA}, Convolutional Neural Network (CNN) \cite{Lecun_CNN}, Sequence-to-sequence (seq2seq) \cite{sutskever2014}, Graph Neural Network (GNN) \cite{Scarselli_GNN_2009}, Autoencoder (AE) \cite{Hinton_AE_2006}, and the Transformer \cite{Vaswani_2017}. 

Architectures are often used in combination. For example, in \cite{An_Tao_Xu_El_Mansori_Chen_2020} a CNN layer is used to perform feature extraction, followed by LSTM layers for prediction. This is a common pattern. CNNs and autoencoders are both widely used for feature extraction in PdM, but they serve different needs. CNNs are typically trained in a supervised way (e.g., fault vs. healthy) to learn features that are directly optimized for a downstream task such as fault classification; they work especially well on time–frequency images (spectrograms) or spatially structured sensor data, and tend to yield highly discriminative features when labeled data are available. Autoencoders, by contrast, are usually trained unsupervised to reconstruct normal operating data, producing latent features that capture the typical behavior of the system; these are especially useful when faults are rare or labels are scarce, and they naturally support anomaly detection via reconstruction error. However, autoencoder features are not guaranteed to be optimal for any specific classification task, while CNN features may fail to generalize to unseen fault modes if the supervised labels do not cover them. In practice, CNNs excel when you have enough labeled data for known faults, whereas autoencoders are preferred for unsupervised health indexing, anomaly detection, and feature pretraining under limited-label conditions.

RNNs, especially LSTMs and GRUs, are used in PdM to model how equipment health evolves over time. They take sequences of measurements (vibration, temperature, pressure, etc.) as input and learn temporal dependencies to: (1) predict Remaining Useful Life (RUL) or time-to-failure as a regression problem, (2) detect anomalies by learning normal temporal patterns and flagging sequences that deviate, and (3) forecast future sensor values or health indicators, which can then feed decision rules for maintenance. RNNs capture short- to medium-range dependencies well but can struggle with very long histories and parallelization. Transformers address this through by using self-attention over the entire sequence, so they handle long-range and cross-sensor interactions more flexibly and are highly parallelizable. This can yield better performance on large, complex datasets, but requires more data and is computationally expensive. In practice, RNNs are often a strong baseline for RUL when data and resources are limited, while Transformers tend to be attractive when there are long sequences, many sensors, and enough data to justify their higher complexity.

Graph Neural Networks (GNNs) offer advantages over sequence models like RNNs and Transformers by explicitly modeling the physical and logical structure of industrial systems. Many assets operate as interconnected networks—machines in a line, components in a turbine, sensors in a an HVAC system—and GNNs can represent these as graphs, passing information along edges to capture how degradation or faults propagate between components. This lets them naturally handle variable numbers of assets and changing topologies, remain invariant to arbitrary ordering of components, and support predictions at multiple levels (component, subsystem, or whole system). As a result, GNNs can exploit relational and structural information that sequence models, which treat data primarily as time-ordered signals, typically ignore or must approximate indirectly.

Table \ref{tab:deep-learning} summarizes common advantages and disadvantages of the above architectures and their applications in PdM. What characterizes all of the above DL architectures is their ``black-box'' nature where the models are not easily interpretable, and the outputs are not explainable. This has several disadvantages in PdM and other high-stakes domains: (1) it is hard to trust or validate the model, especially when its recommendation conflicts with expert judgment or carries significant cost (e.g., shutting down a production line); (2) debugging and improving the model is difficult, because we cannot easily see whether it relies on spurious correlations, sensor faults, or dataset biases; (3) regulatory, safety, and contractual requirements may demand traceable rationales for decisions; and (4) engineers cannot easily translate learned patterns into human knowledge that can be reused, audited, or combined with existing rules and physical models. 

Post-hoc interpretability and explainability techniques address some of these issues. Interpretability aims for models whose structure and parameters have clear meaning (e.g., rules, physical parameters, or logical predicates), while explainability focuses on generating understandable reasons for individual predictions (e.g., which sensors and time windows drove an RUL estimate). Together, they aim to enable trust and adoption by domain experts, support root-cause analysis and model debugging, help detect data or concept drift, and facilitate compliance with safety constraints. To that end, a variety of techniques have been proposed, which are outlined in Table \ref{tab:xai}. A detailed analysis of Explainable AI (XAI) techniques is beyond the scope of this work, thus we refer the interested reader to \cite{Vilone_Longo_2020,Vilone_Longo_2021}.

\begin{table}[ht]
    \footnotesize
    \renewcommand{\arraystretch}{1.5}
    \begin{tabular}{%
    >{\raggedright\arraybackslash}p{4.5cm}%
    >{\raggedright\arraybackslash}p{6.5cm}%
    }
        \toprule
        Technique & Limitations \\
        \midrule
        \textbf{Feature importance} (e.g., permutation importance, tree-based importance) to rank critical sensors and variables. & Often global and coarse; can miss interactions, be unstable across runs, and be misleading when features are correlated or when time order matters. \\
        \textbf{Partial Dependence Plots (PDP) / ICE plots} to show how a feature affects RUL or failure risk. & Assume partial independence of features, which is rarely true in multivariate sensor data; become hard to interpret in high dimensions and do not capture temporal causality.\\
        \textbf{LIME} (Local Interpretable Model-agnostic Explanations) for local, instance-level explanations of black-box predictions. & Explanations depend heavily on how local neighborhoods are sampled; can be unstable (different runs give different explanations) and may not reflect the model’s true global behavior. \\
        \textbf{SHAP} (SHapley Additive exPlanations) to quantify each feature’s contribution to a specific prediction over time. & Computationally expensive for large models or long time series; assumes additive feature contributions, which can oversimplify complex dependencies; results can be difficult to interpret for many features and time-steps. \\
        \textbf{Saliency maps / Grad-CAM / integrated gradients} for CNN-based models on vibration spectrograms or other signal images. & Sensitive to noise and model initialization; different methods can yield different heatmaps; highlight where the model looks but not why in human terms. \\
        \textbf{Attention weights visualization} for RNN/Transformer models to highlight influential time steps and sensors. & ``Attention $\neq$ explanation''; high attention weights do not always correspond to true causal importance; multi-head and multi-layer attention are hard to interpret clearly.\\
        \textbf{Surrogate models} (e.g., decision trees or rule lists) approximating a complex PdM model for global understanding. & Only approximate the original model; risk of over-simplification or misrepresentation, especially in regions with complex decision boundaries. \\
        \textbf{Prototype and counterfactual examples} (similar past trajectories, or ``what-if'' changes to sensor values) to illustrate model behavior. & Choosing meaningful prototypes and realistic counterfactuals is non-trivial; generated examples may be implausible given real system dynamics or constraints. \\
        \textbf{Rule extraction} to express learned patterns as human-readable maintenance rules. & Often approximate, incomplete, and costly to compute; rules can become numerous, inconsistent, or too complex for humans to reliably inspect and maintain.\\
        \bottomrule
    \end{tabular}
    \caption{Post-hoc explainability techniques.}
    \label{tab:xai}
\end{table}


\begin{table}[ht]
    \centering

    \footnotesize
    \renewcommand{\arraystretch}{1.5}
    \begin{tabular}{%
    >{\raggedright\arraybackslash}p{1.5cm}%
    >{\raggedright\arraybackslash}p{4cm}%
    >{\raggedright\arraybackslash}p{5.5cm}%
    }
    \toprule
    Architecture & Use cases & Advantages and Disadvantages \\
    \midrule
    CNN & Fault diagnosis from vibration and acoustic signals; classification from time–frequency images (e.g., spectrograms); sensor-based health state recognition. & 
    Excels at local pattern extraction; handles high‑dimensional signals and images; automatic feature learning; relatively efficient inference. 
    \newline \newline 
    Needs substantial labeled data; less suited to very long temporal dependencies (unless combined with RNN/Transformer); can be hard to interpret. 
    \\
    Seq2Seq (RNN/ LSTM/ GRU) & Remaining Useful Life (RUL) prediction from multivariate time series; sequence-based anomaly detection; temporal degradation modeling. & 
    Naturally models temporal dependencies; works directly on raw or lightly processed time series; good for sequence-to-one and sequence-to-sequence tasks. 
    \newline \newline 
    Training can be slow and unstable; struggles with very long sequences vs. Transformers; harder to parallelize; risk of vanishing/exploding gradients. \\
    
    AE & Unsupervised anomaly detection on sensor data; health index learning; dimensionality reduction and feature extraction for downstream PdM models. & 
    Learns compact representations without labels; well suited to rare-fault scenarios; flexible (can be CNN-, RNN-, or MLP-based); useful for denoising. 
    \newline \newline 
    Threshold selection for anomalies is non-trivial; may reconstruct faults too well if trained on mixed data; representations can be hard to interpret. \\
    
    GNN & Modeling systems with interacting components (e.g., production lines, power grids); fault localization on networks; RUL prediction for asset fleets with relational structure. & 
    Captures topology and inter-component dependencies; naturally handles variable-size systems and graph-structured assets; can share information across similar components. 
    \newline \newline 
    Graph construction and feature design are non-trivial; scalability can be challenging for very large graphs; still relatively new in PdM with limited tooling and benchmarks \\
    
    Transformer & Long-horizon RUL prediction from multivariate time series; forecasting-based PdM; multivariate sequence anomaly detection; sensor fusion across many channels. & 
    Handles long-range dependencies well; highly parallelizable; flexible input modalities; can model complex temporal and cross-sensor interactions. 
    \newline \newline 
    Data‑hungry and computationally expensive; prone to overfitting in small-data industrial settings; hyperparameter tuning is complex. \\
    \bottomrule
    \end{tabular}
    \caption{Common Deep Learning architectures and characteristics in PdM.}
    \label{tab:deep-learning}
\end{table}

\subsection{Hybrid} \label{sec:hybrid}
\begin{figure}[ht]
    \centering
    \includegraphics[width=1\textwidth]{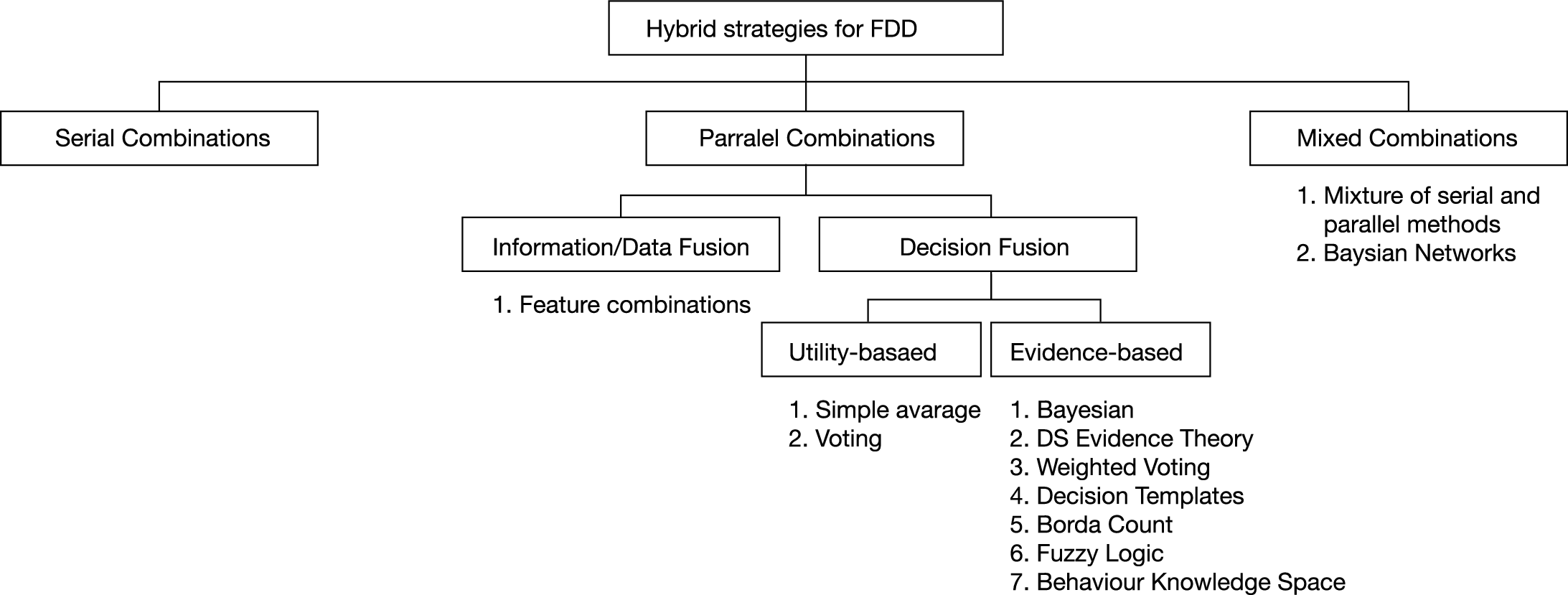}
    \caption{Combination strategies for hybrid FDD identified by \cite{Wilhelm_2021}}
    \label{fig:hybrid-fdd}
\end{figure}

\textit{Hybrid} methods are characterized by the use of multiple models for a single task. This is considered the current leading trend in PdM research. Models can be combined in series (output of one is input to another), in parallel (outputs are fused), or in an embedded structure \cite{Jimenez_2020}\endnote{In contrast to \cite{Jimenez_2020}, we classify the embedded type as \textit{nested} neuro-symbolic rather than hybrid.}. Similarly, \cite{Wilhelm_2021} divide hybrid approaches into serial and parallel and also provide examples of the combination of both. An overview of combination strategies for Fault Detection and Diagnosis (FDD), is reproduced from \cite{Wilhelm_2021} in Figure \ref{fig:hybrid-fdd}. 

\subsubsection{In-series:} \label{sec:in-series}
Several examples of in-series implementations are given next.
\cite{Villalobos_2021} propose a ``forecaster-analyzer'' system where the time-series data is forecast using an LSTM, the output of which is used to train a Residual Neural Network (ResNet) analyzer to predict alarms.

In \cite{Stiefelmaier_2024}, the authors utilize the \textit{reconstruction method} followed by model-based fault detection and isolation (i.e., classification). The authors identify a core problem in model-based fault diagnosis as being the lack of robustness to uncertainties and errors in the underlying model. To address this issue, the authors train an autoencoder (\textit{reconstruction}), in an unsupervised fashion, to suppress the effects of (physical) model errors in parity space residuals. Since the presence of a fault as modeled above results in a change in the residual’s mean, the fault detection problem amounts to detecting a change in the mean of a sequence. The results of this hybrid-AE method are superior to model-based fault diagnosis, but on par with a similarly constructed hybrid-PCA method.

\subsubsection{In-parallel:}
Examples of in-parallel implementations follow. In \cite{Lee_Kim_2025}, machine learning and statistical features are combined in parallel into a single feature vector for fault detection in rotating machinery. A 1-D CNN is trained on pseudo labels extracted from the data using rotating speed information (i.e. domain knowledge of the underlying physics) in order to generate vector representations of the data samples. Features are extracted using vibration statistics such as mean, peak, rms, crest factor, skewness, and kurtosis, to form a 6 dimensional vector. These two vectors are concatenated and used to fit a Local Outlier Factor (LOF) density based k-nearest-neighbor (KNN) algorithm for fault detection, where outliers are classified as faults. The authors report precision and recall scores well above single model state-of-the-art approaches on several rotating machinery data sets (UOS \cite{Lee_Kim_Kim_2024}, HUST \cite{Zhao_Zio_Shen_2024}, and XJTU-SY \cite{ Wang_Lei_2020}). Notably, the hybrid features result in precision and recall scores above 95\% under varying operating conditions, and where the training set does not contain labels. 

\subsection{Modeling Approaches Summary}
Physics-based models offer strong interpretability and extrapolation grounded in physical laws but demand detailed system knowledge, are costly to develop, and can be hard to maintain. Knowledge-based models capture expert reasoning and are relatively easy to understand and modify, yet they depend on scarce expert input, struggle with coverage, and may not adapt well to changing conditions. Data-driven models scale well, can capture complex patterns, and often deliver high predictive accuracy, but they require large, representative datasets, may be opaque (``black boxes''), and often generalize poorly to unseen conditions. Hybrid methods combine multiple models and techniques in order to leverage the strengths of each. It is worth noting that most recent work in the literature combines multiple techniques, even if it does not explicitly label them as hybrid. 

Most recently, Neuro-symbolic (NESY) methods have gained attention; they differ from traditional hybrid approaches mainly in the depth of integration between symbolic and subsymbolic components which we will discuss in detail. Although there are currently few NESY applications in PdM, we argue that they offer a promising direction for achieving more robust, inherently interpretable, and efficient systems. Accordingly, the next section is dedicated exclusively to NESY.

\section{Neuro-symbolic AI (NESY)} \label{sec:neuro-symbolic}

\begin{figure}[ht]
    \centering
    \includegraphics[width=1\linewidth]{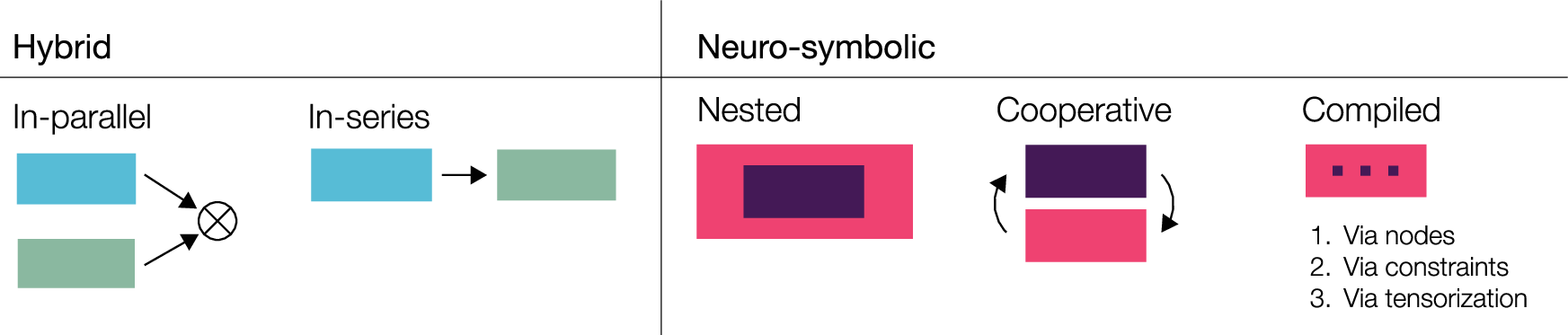}
    \caption{Hybrid vs. NESY architectures}
    \label{fig:hyb-nesy}
\end{figure}

In the Neuro-symbolic paradigm, symbolic knowledge and neural networks are integrated, or tightly coupled, to produce a single architecture. Ideally, such an architecture not only learns from data and symbolic knowledge, but the resulting model can be used for reasoning beyond the training data distribution (although, this is not always the case). 

Several taxonomies have been proposed. One of the most cited is the Kautz taxonomy introduced by Henry Kautz at the Thirty-fourth AAAI Conference on Artificial Intelligence \cite{Kautz_2022}. Kautz categorizes NESY architectures into six types, roughly in order of the level of integration as follows:

\begin{enumerate}
    \item \textbf{Symbolic Neuro symbolic} is currently the deep learning SOP (standard operating procedure) for natural language processing.
    \item The \textbf{Symbolic[Neuro]} architecture employs a Neural pattern recognition subroutine within a symbolic problem solver.
    \item In a \textbf{Neuro | Symbolic} system, a neural network converts nonsymbolic input, such as the pixels of an image, into a symbolic data structure, which is then processed by a symbolic reasoning system.
    \item The \textbf{Neuro: Symbolic → Neuro} approach uses the SOP architecture but with a special training regime based on symbolic rules.
    \item A \textbf{Neuro\_{Symbolic}} architecture transforms symbolic rules into templates for structures within the neural network.
    \item \textbf{Neuro[Symbolic]} architecture where a symbolic reasoning engine inside a neural engine, with the goal of enabling super-neuro and combinatorial reasoning. This, Kautz states, has the greatest potential to combine the strengths of logic-based and neural-based AI.
\end{enumerate}

\begin{figure}
    \includegraphics[width=1\linewidth]{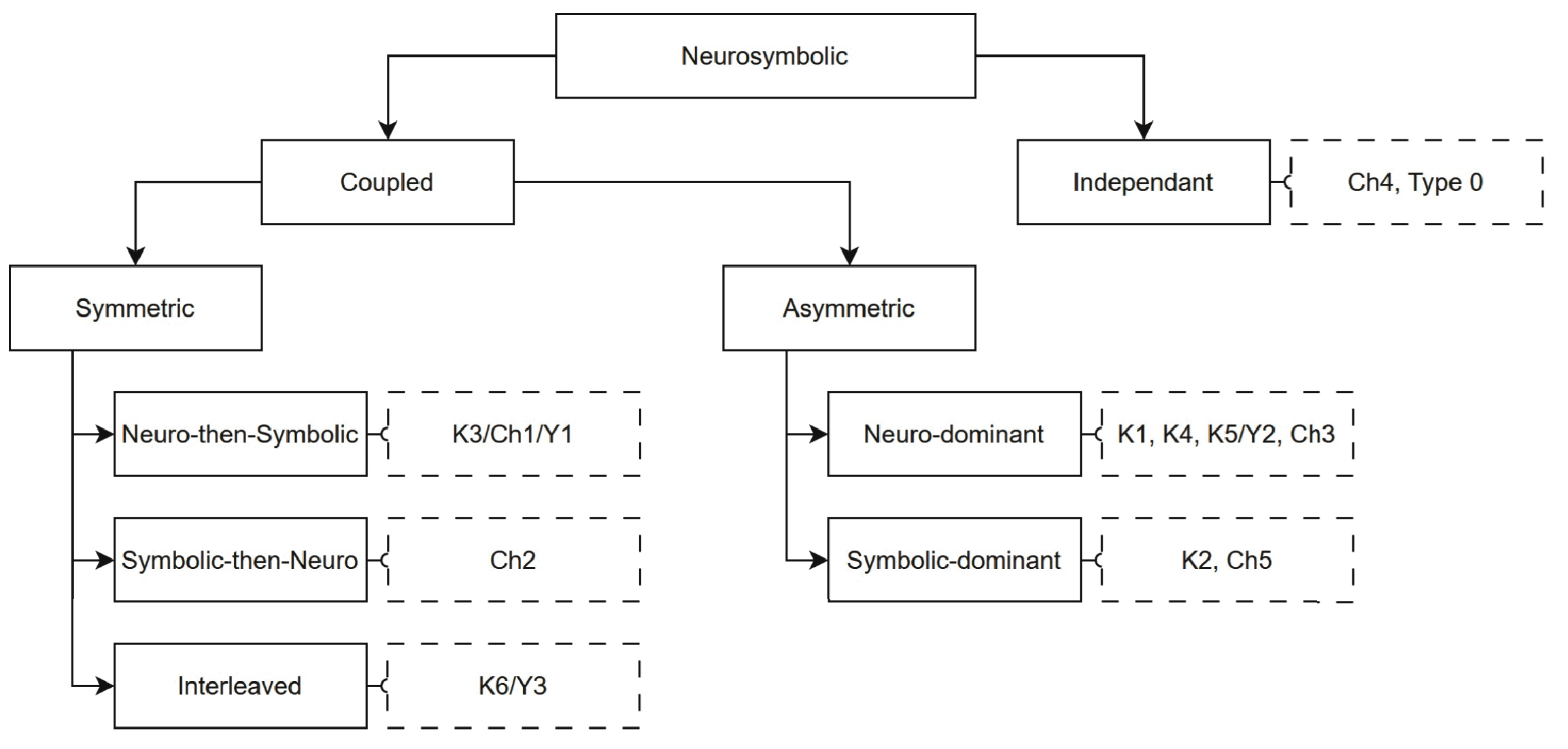}
    \caption{NESY taxonomy reproduced from \cite{Pacheco_2025}. ``\textit{Types from the taxonomies of Kautz, Chaudhuri, and Yu are abbreviated using the author’s initial followed by their type number. Approaches separated by slashes denote equivalence; those separated by commas represent distinct types grouped within the same category.}''}
    \label{fig:pacheco}
\end{figure}

\cite{Yu_Yang_Liu_Wang_Pan_2023} categorize neuro-symbolic systems into three types: \textbf{learning-for-reasoning}, \textbf{reasoning-for-learning}, and \textbf{learning-reasoning}, while \cite{Pacheco_2025} propose rearranging the above mentioned taxonomies from the perspective of module dominance and order of operations which ``aim to capture both the structural dependencies and the interaction dynamics between neural and symbolic components'' - Figure \ref{fig:pacheco}. While a single agreed upon taxonomy of NESY has not been adopted yet, based on our prior work in \cite{Hamilton_Nayak_Bozic_Longo_2024}, we categorize NESY architectures into three major types: \textit{Nested}, \textit{Cooperative}, and \textit{Compiled} - Figure \ref{fig:hyb-nesy}. We believe this terminology is both intuitive and flexible enough to cover most NESY designs. These are described in the following subsections with examples from the PdM domain where available.



\vphantom{------------
9.1. Dynamic neuro-symbolic systems \cite{Nawaz_Anees_2025}
Dynamic neuro-symbolic systems signify the further advancement in hybrid AI models by facilitating real-time adaptation of AI to evolving settings (Bhuyan et al. (2024)). Conventional symbolic systems depend on fixed rules, but neural networks may evolve through data-driven learning. Dynamic neuro-symbolic systems integrate these advantages by enabling the symbolic component to modify or develop regulations in response to new information the neural networks achieve. This flexibility is essential for systems functioning in non-stationary contexts, such as autonomous agents or industrial robots, where norms and patterns fluctuate constantly (Lu et al. (2024)). A dynamic neuro-symbolic system may modify its symbolic reasoning framework in response to new sensory inputs from the neural network. These systems are being investigated for continuous learning tasks and can mitigate the inflexibility of conventional symbolic methods.
nesy ai facilitates real-time decision-making among interconnected devices (Lu et al. (2024)). IoT systems provide extensive data from sensors, cameras, and many sources. Neural networks analyze unstructured data proficiently, whereas symbolic reasoning facilitates decision-making based on logical norms and constraints, like safety laws and energy consumption thresholds (Zeng et al. (2023)). In smart cities, neuro-symbolic systems can evaluate real-time traffic data and implement symbolic rules on traffic management to make adaptive decisions, such as rerouting cars to alleviate congestion. The difficulty with IoT systems is the need for rapid, dependable decisions in settings with limited processing resources (Gubbi et al. (2013) and Lee et al. (2015)). In conjunction with lightweight neuro-symbolic architectures, edge computing is being investigated as a solution that facilitates decision-making near the data source and diminishes delays linked to cloud-based processing.\cite{Nawaz_Anees_2025}
-----------}

\subsection{Nested}
This paradigm involves either a neural module inside a symbolic system, or vs, a symbolic module inside a neural system. For example, a symbolic system which performs mathematical computations contains internal neural modules which are utilized to recognize numbers and symbols from images. AlphaGo \cite{Silver_Huang_2016} is the example given by Henry Kautz \cite{Kautz_2022}, where the external symbolic system is a Monte Carlo Tree Search \cite{Coulom_2007} with internal neural state estimators nominating next states. Most robots and autonomous vehicles are nested, or in Kautz's terminology, \textit{Symbolic[Neuro]} systems. There are few to no examples of nested systems where the internal module is symbolic with an external neural engine or \textit{Neuro[Symbolic]}. Kautz likens \textit{Neuro[Symbolic]} systems to Kahneman's System 1/System 2 \cite{kahneman2011thinking} human level thinking, where System 1 is fast and associative akin to an ANN, and System 2 is slow and deliberative, akin to symbolic reasoning. The neural system generates symbolic representations of the given problem and triggers a symbolic system as needed.

\subsection{Cooperative}
This architecture is made up of 2 or more neural and symbolic modules which learn from each other in an iterative fashion. The errors from one module inform the subsequent module. Here the modules are co-routines, in contrast to subroutines in the nested type. It should be noted that in both cases the modules are dependent on each other's output, which distinguishes these systems from \textit{hybrid} ones where the modules act independently. An example of a cooperative architecture is \cite{Mao2019}'s Neuro-Symbolic Concept Learner (NS-CL) which jointly learns visual concepts, words, and semantic parsing of sentences without any explicit annotations. Given an input image, the visual perception module detects objects in the scene and extracts a deep, latent representation for each of them. The semantic parsing module translates an input question in natural language into an executable program given a domain specific language (DSL). The generated programs have a hierarchical structure of symbolic, functional modules, each fulfilling a specific operation over the scene representation. The explicit program semantics enjoys compositionality, interpretability, and generalizability. Reinforcement Learning (RL) \cite{Kaelbling_Littman_Moore_1996} is commonly applied in Cooperative systems.

\subsection{Compiled}
Whereas the above NESY types contain separate symbolic and neural modules that interoperate in different ways, here symbolic knowledge is fully integrated, or ``compiled'', into the neural network itself. These can be further categorized into architectures where knowledge is integrated in the network nodes (NESY-CN), ones where knowledge is integrated as constraints inside the loss function (NESY-CL) and thirdly where symbolic knowledge is embedded in vector space, or what is known as \textit{tensorization}.  This makes the model’s behavior more consistent with domain knowledge while remaining fully differentiable and trainable with standard optimization methods.

\begin{wrapfigure}{r}{0.5\textwidth}
    \includegraphics[width=.95\linewidth]{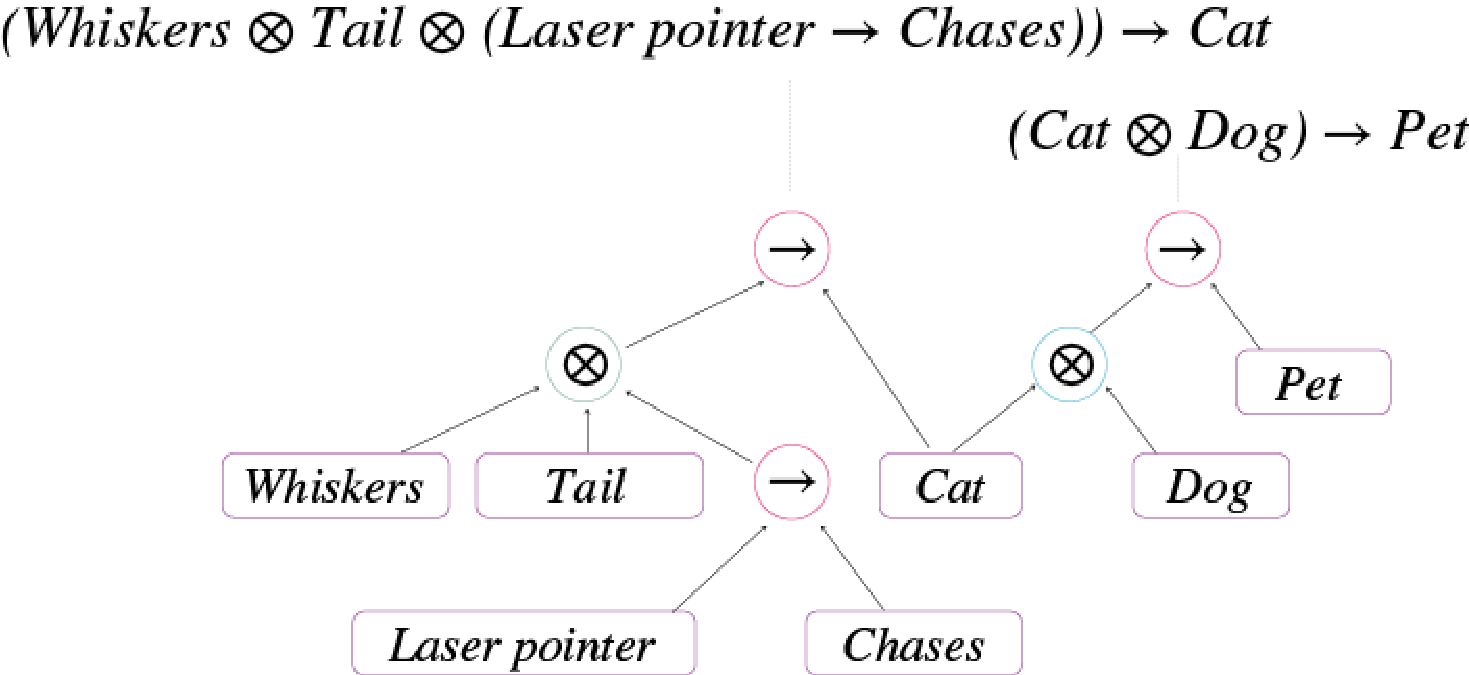}
    \caption{Logic Neural Network (LNN) - reproduced from \cite{Riegel_Gray_2020}.}
    \label{fig:lnn}
\end{wrapfigure}

\subsubsection{Compiled into network nodes (NESY-CN):}
What makes NESY-CN systems particularly attractive is their potential for inherent interpretability. In NESY-CN, knowledge comes in the form of rules (i.e., $IF \ A, \ THEN \ B$), logic expressions (i.e., $A \ \land \ B$) or analytic expressions (i.e., $Ax^2 + Bx + C$) which are directly related to the system dynamics, which in turn is a step in the direction of root cause analysis, a simultaneously challenging and critical component of industrial maintenance. 

The foundation of NESY-CN is the Logic Neural Network (LNN) introduced by \cite{Riegel_Gray_2020}. An LNN is a neural model that represents logical rules directly inside its structure. It builds layers that correspond to logical operations like AND, OR, and NOT, and assigns weights to rules to show how confident the model is in them. Inputs are simple facts, and as they pass through these logical layers, the network computes how true different conclusions are. The whole system is trained via gradient-descent, so it can adjust both how it interprets facts and how strongly it trusts each rule, combining data-driven learning with explicit, human-readable logic. Figure \ref{fig:lnn} illustrates the LNN architecture.

\begin{wrapfigure}{l}{0.5\textwidth}
    \includegraphics[width=.95\linewidth]{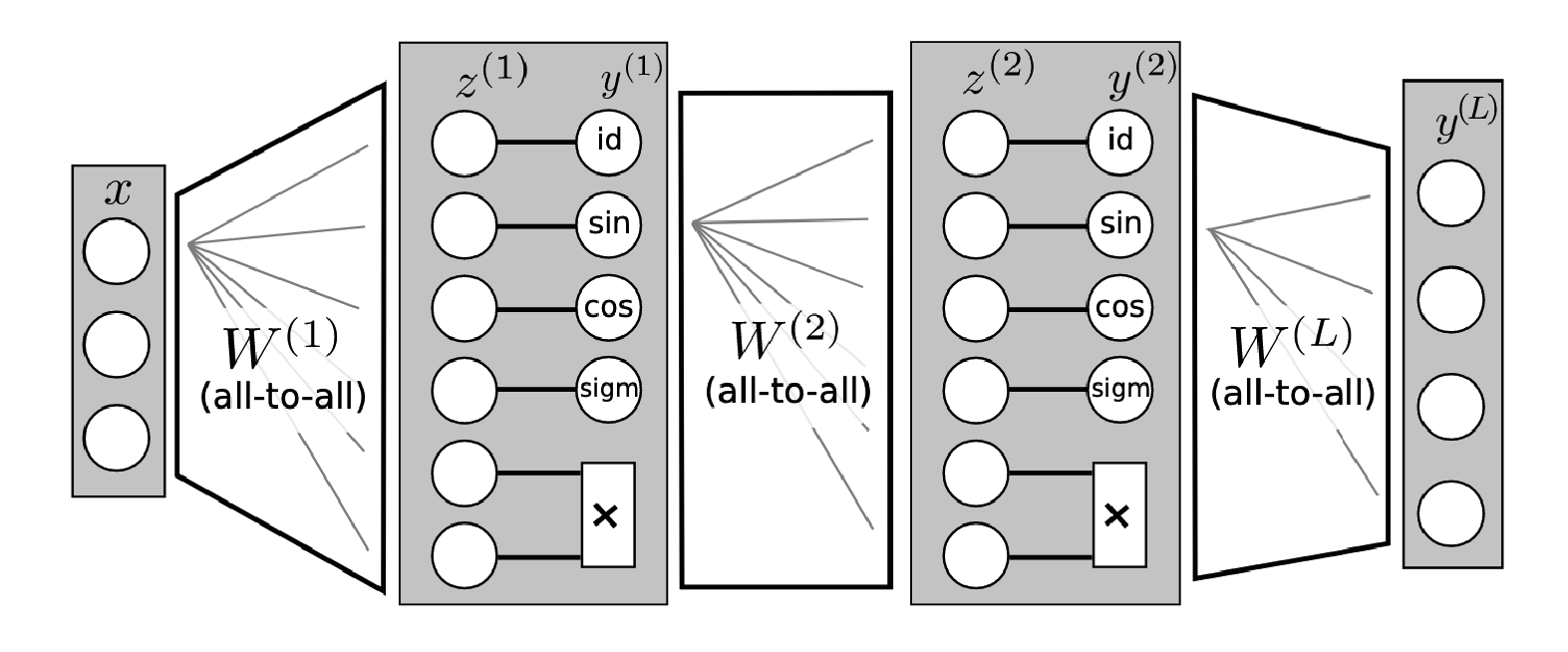}
    \caption{The Equation Learner (EQL) architecture reproduced from \cite{MartiusL16}.}
    \label{fig:eql}
\end{wrapfigure}

\textbf{The Equation Learner (EQL)} \cite{MartiusL16} adapts a Multi-Layer Perceptron (MLP) by using predefined symbolic functions as activations and then sparsifying connections to extract compact analytic expressions from data. However, it struggles to sparsify fully connected networks effectively — L1 regularization often cannot reduce connections without harming accuracy — leading to poor fits. Additionally, its fixed architecture cannot adapt to task complexity, causing either overly complex expressions when too large or degraded fitting when too small. The EQL architecture is depicted in Figure \ref{fig:eql}.


Similarly, \textbf{Neural Reasoning Network (NRN)} \cite{Carrow_Gray_2024}, inspired by Logical Neural Networks (LNN) \cite{Riegel_Gray_2020}, consists of layers of sparsely connected nodes whose activation functions are logical operators. These operators are implemented using Modified Weighted Łukasiewicz Logic formulas (Table \ref{tab:NRN-functions}) which are conducive to gradient descent. The network is trained using reinforcement learning, culminating in a tree structure. This structure is simplified into a set of rules to be used for inference. The resulting model is both parsimonious and easily interpretable by humans. Unlike the EQL above, the NRN architecture is not fixed but learned from the data. 


\begin{table}[ht]
    \centering
    \begin{tabular}{ll}
    \toprule
        Operator & Formula \\
        \midrule
        Conjunction (AND) & 
        $
        f\left( 1 - \sum_{j} w_j(1-x_j)\right)
        $\\
        Disjunction (OR) & 
        $
        f\left(\sum_{j} w_jx_j\right)
        $\\
        \bottomrule
    \end{tabular}
    \caption{Modified Weighted Łukasiewicz Logic activation functions \cite{Carrow_Gray_2024}. Where $w_j > 0$ and $f$ clamps the result in the range $[0,1]$.}
    \label{tab:NRN-functions}
\end{table}

\begin{figure}
    \includegraphics[width=1\linewidth]{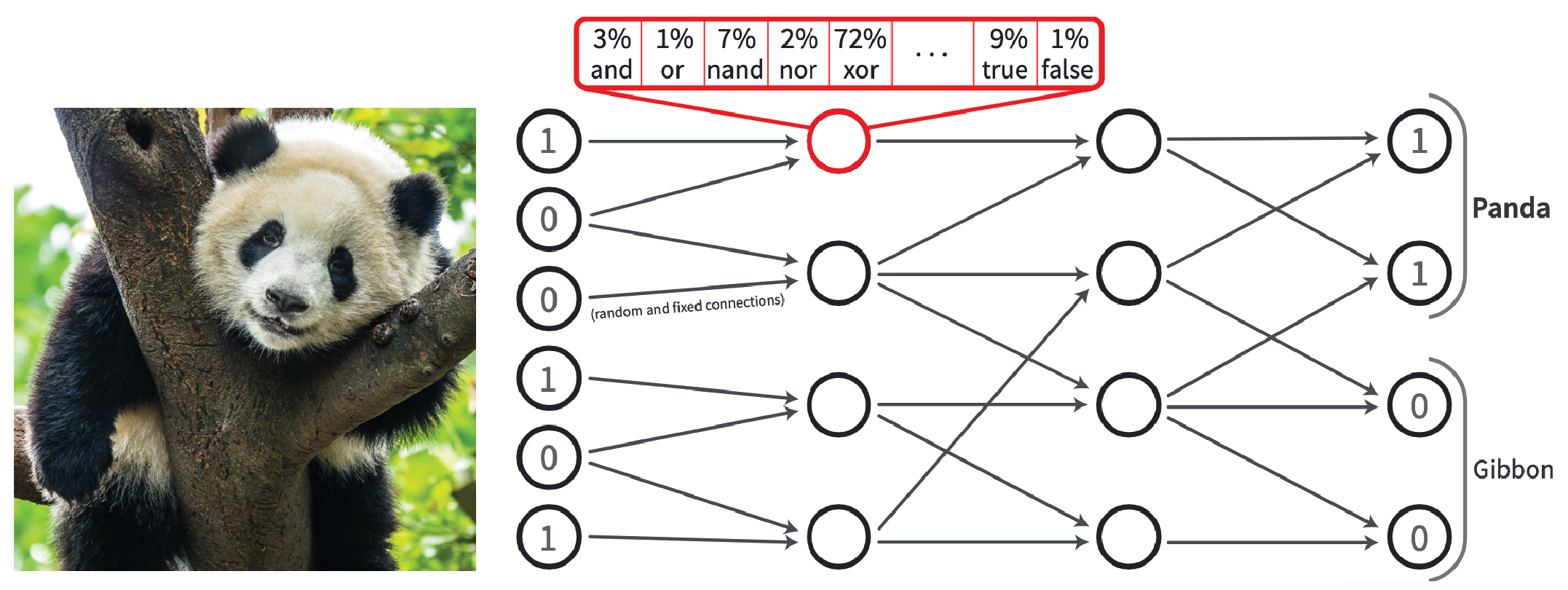}
    \caption{DiffLogicNet. During training, the distribution of logic gates is learned for each neuron, and, during inference, the most likely operator is used.}
    \label{fig:diffLogic}
\end{figure}

\cite{Petersen_2022} propose \textbf{Deep Differentiable Logic Gate Networks (DiffLogicNet)}\endnote{\url{https://github.com/Felix-Petersen/difflogic}}, an approach for gradient-based training of logic gate networks. Logic gate networks are based on binary logic gates, such as $AND$ and $XOR$. For training they use real-valued logic and learn which logic gate to use at each neuron. Specifically, for each neuron, a probability distribution over logic gates is learned. The goal is to learn which logic gate operators are present at each neuron. After training, the resulting network is discretized to a (hard) logic gate network by choosing the logic gate with the highest probability. As the (hard) logic gate network comprises logic gates only, it can be executed very fast. Additionally, as the logic gates are binary, every neuron / logic gate has only 2 inputs, and the networks are extremely sparse. Figure \ref{fig:diffLogic} illustrates the DiffLogic architecture.

Specific to the AFDD domain, \cite{li_deep_2024} propose \textbf{Deep Expert Network}, ``a fully interpretable IFD [Intelligent Fault Diagnosis] approach that incorporates expert knowledge using neuro-symbolic AI. This study specifically centers on examining the primary bearing fault types associated with rotating machinery.'' Here, ``expert knowledge'' are certain statistical features such as Hadamard product, Fourier Transform (FT), Moving Average (MA), and kurtosis, among others, rather than rules defined by human experts. These features are passed to a fuzzy logic layer (which constitutes the compiled part of the system) and the network is learned via gradient descent. The network is then pruned to reduce its size resulting in an interpretable tree-like structure where the nodes are features and logical operators. This provides a ``decision route'' from the data to the prediction. Accuracy is slightly higher than the reported black-box baselines. 


\textbf{Hamiltonian Neural Networks (HNN)} \cite{Greydanus_Dzamba_Yosinski_2019} embed predefined physics (in the form of differential equations) in the network structure. By learning the system's total energy, they guarantee conservation laws are respected, which leads to long-term stability for conservative systems. The output of an HNN is a single value, the Hamiltonian, with perfectly energy-conserving dynamics. Because HNNs offer guaranteed behaviour, they are highly relevant for industrial and safety-critical applications such as aerospace, nuclear power, or advanced robotics, where model reliability and predictable behaviour are paramount. Although HNNs are limited to modeling closed systems, extensions like \textbf{Port-Hamiltonian Neural Networks (PHNNs)} \cite{Desai_Mattheakis_Sondak_Protopapas_Roberts_2021} and \textbf{Output Error Hamiltonian Neural Network (OE-HNN)} \cite{Moradi_Jaensson_Tóth_Schoukens_2023}  are actively being developed to overcome this limitation by accounting for phenomena such as friction, damping, or other external forces and noise. Such networks have potential in PdM for certain types of signals such as from rotating machinery (motors, pumps, fans, generators, or compressors). 

    
    
\subsubsection{Compiled into loss (NESY-CL):} Here, logical knowledge is integrated directly into the loss function so that learning is driven not only by data labels but also by logical constraints. During training, violations of rules incur penalties, encouraging the network to find parameters that satisfy both empirical data and the specified logic.

\textbf{Physics Informed Neural Networks (PINN)}, introduced by \cite{Raissi_Perdikaris_Karniadakis_2019}, are neural networks where known physical formulas in the form of partial differential equations (PDEs) are included as soft constraints in the loss function (similar to how other well known regularization terms such as L1 and L2 regularization are included). The parameters of a physical constraint can be predefined or learned. The network thus learns from both the data and the knowledge of the physics being modeled. The parameters of the physics terms can also be learned. The output is a physically-consistent approximated solution. This approach has potential in PdM where the systems are governed not only by external, environmental, or temporal factors, but also by their underlying physics. 

\begin{figure}[ht]
    \centering
    \includegraphics[width=0.75\linewidth]{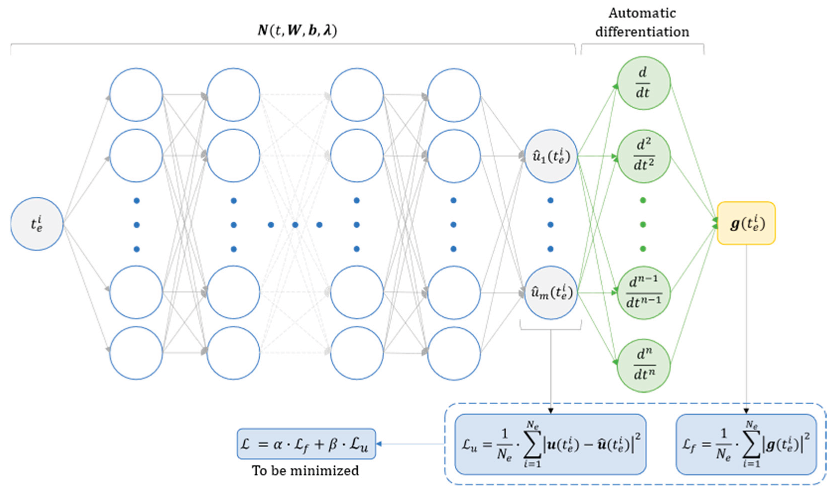}
    \caption{PINN architecture. Note that the hyperparameters of the PINN to be optimized are not only the weights $W$ and biases $b$ but also, and significantly, the parameter vector $\lambda$ describing the system health state \cite{Parziale_2023}.}
    \label{fig:pinn}
\end{figure}

\cite{Parziale_2023} show that a physics informed CNN has better overall accuracy than a CNN alone, for estimating unknown parameters characterizing the health state of rotating shaft systems. Figure \ref{fig:pinn} illustrates the network architecture of a PINN which includes the loss function containing the PDEs.

\cite{Liao_Chen_Wen_Zhao_2023} propose a PINN model with an attention mechanism, AttnPINN, which out-performs previous SOTA methods on NASA’s turbofan engine degradation C-MAPSS dataset and has the additional benefit of interpretability. 

\cite{Ren_Xu_2023} propose \textbf{Thermodynamic-law-integrated Autoencoder (TLI-Autoencoder)} where the laws of thermodynamics are integrated into the loss function of an autoencoder. They demonstrate a reduction on the false positive rate of 77.4\% and an increase of 14\% on the fault detection rate over a standard autoencoder trained on data collected from a real chiller plant. Unlike the previous examples, training is performed in an unsupervised setting. The authors note that a single sensor could be governed by multiple thermodynamic laws. Therefore, the thermodynamic-law-induced loss should be a summation of the losses induced by the laws governing the sensors. 

\cite{leung_backpropagation_2023} propose several avenues for leveraging  Signal Temporal Logic (STL) in neural networks by representing STL formulas as computational graphs. In particular, the authors provide a NESY-CL use case whereby STL formulas are encoded in the loss function of a Variational Auto Encoder (VAE) for more robust time-series forecasting. The authors demonstrate that ``infusing desirable behaviors into the model can improve long-term prediction performance.'' This offers a promising direction for RUL prediction. For example, STL can encode temporal patterns such as: \textit{the temperature of the room will reach 25 degrees within the next 10 minutes and will stay above 22 degrees for the next hour.} \cite{Stiefelmaier_2024}. Such STL formulas can be turned into differentiable loss terms: when a neural RUL predictor or anomaly detector proposes trajectories that violate these temporal rules, it is penalized during training, thereby steering the model toward physically and operationally plausible behaviors even with sparse labels. The residuals between the model predictions and actual signals can then be utilized to detect anomalies, or in other words, to detect faults.



\subsubsection{Compiled into tensors (NESY-CT):} Some of the earliest compiled NESY formulations include \textbf{Logic Tensor Networks (LTN)}, and \textbf{Tensor Product Representations (TPR)}, commonly grouped under the moniker of \textit{tensorization}. Both aim to bridge symbolic reasoning with continuous vector spaces by encoding symbols and their relations in tensor-based structures that can be manipulated by neural methods. 

\begin{figure}[ht]
\centering
\begin{subfigure}{0.45\textwidth}
    \includegraphics[width=\textwidth]{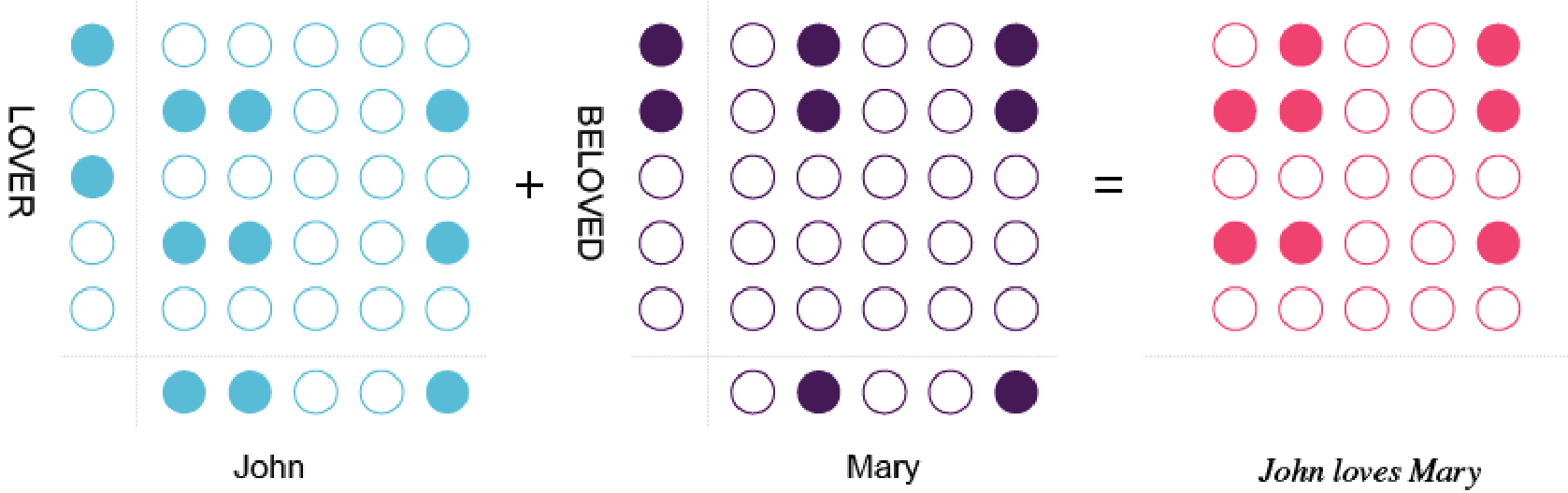}
    \caption{Tensor Product Representation}
    \label{fig:tpr}
\end{subfigure}
\hfill
\begin{subfigure}{0.45\textwidth}
    \includegraphics[width=\textwidth]{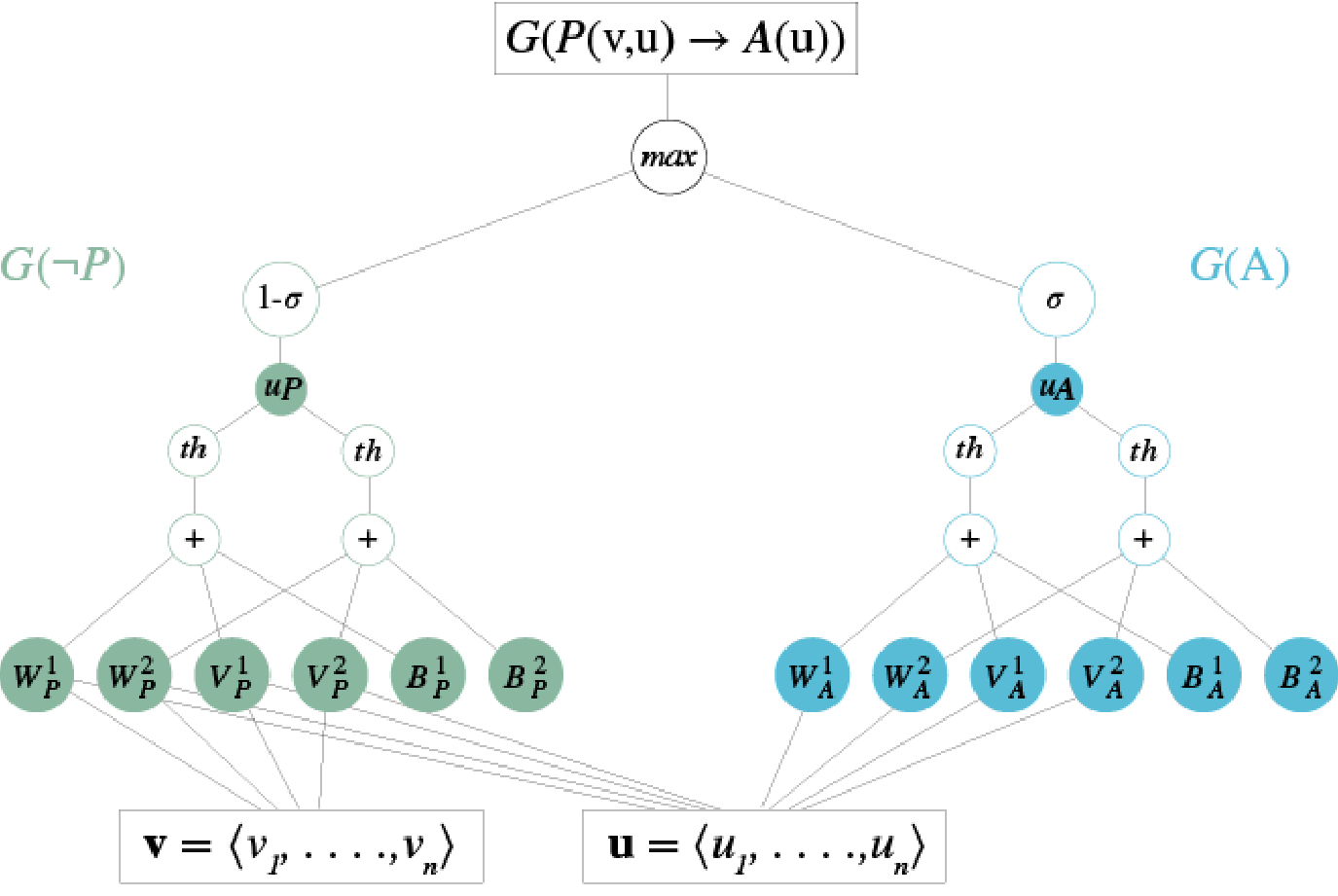}
    \caption{Logic Tensor Network}
    \label{fig:ltn}
\end{subfigure}
        
\caption{Tensorization}
\label{fig:tensorization}
\end{figure}

TPRs aim to give neural networks a principled way to represent and manipulate symbolic structures—such as trees, sequences, and variable bindings—using continuous vectors. By encoding “fillers” (symbols) and ``roles'' (positions or relational slots) as vectors and ``binding'' them via tensor products, TPRs provide a mathematically precise scheme for variable binding and compositionality in distributed representations. The goal is to let neural systems handle structured, symbolic information (like language or logic) in a way that is both expressive and compatible with standard linear algebra and learning methods, as illustrated in Figure \ref{fig:tpr}.

LTNs aim to combine the strengths of neural networks and first-order logic in a single, trainable framework, illustrated in Figure \ref{fig:ltn}. They seek to represent objects as vectors and predicates as differentiable functions so that logical formulas become real-valued constraints whose satisfaction can be optimized with gradient descent. This allows LTNs to jointly learn from data and logical knowledge, perform approximate logical reasoning, handle uncertainty and partial truth, and remain interpretable in terms of explicit rules and predicates, rather than acting as a black box.

LTNs suffer from  combinatorial explosion and poor scalability on large domains or deep theories. Similarly, Tensor products cause dimensionality and memory blowup as structure depth, arity, or vocabulary size grows. Systems based on \textit{tensorization} aim to address these limitations while preserving benefits such as interpretability, determinism, and reasoning.

An example of \textit{tensorization} is the \textbf{Deep Knowledge Augmented Belief Network (DKABN)} \cite{DKABN_2025}. DKABN integrates prior knowledge in the form of propositional logic rules which are embedded in the network. Performance was evaluated using a public bearing dataset (vibration signals). DKABN achieves a favorable trade-off among model interpretability, computational overhead, and fault recognition performance for diagnostic decision-making.


\subsection{NESY Summary}

NESY architectures can address several concrete pain points in PdM on time‑series data such as data scarcity, false alarms, poor generalization, and lack of interpretability, by design. \textit{Nested} and \textit{cooperative} systems mainly add flexible combinations of neural perception with symbolic planning or querying, but the key benefits come from \textit{compiled} approaches. Node‑level compiled models (NESY‑CN) such as Logic Neural Networks, Neural Reasoning Networks, DiffLogicNet, Deep Expert Networks, and Hamiltonian NNs turn rules, fuzzy/real‑valued logic, expert features, or physics equations directly into network operations; this yields (i) inherently interpretable decision paths (often pruned into human‑readable trees or equations), (ii) rule‑weighted reasoning that can be inspected and edited, and (iii) predictions that respect known physical structure (e.g., energy conservation in HNNs). Loss‑level compiled models (NESY‑CL), including PINNs, thermodynamic‑law‑integrated autoencoders, and STL‑regularized VAEs, use physics, thermodynamic constraints, or temporal logic over sensor trajectories as differentiable penalties. These architectures have empirically reduced false positives, improved RUL accuracy, and produced more physically plausible forecasts, especially under noisy or limited labels. \textit{Tensorized} approaches (NESY‑CT), such as Logic Tensor Networks and knowledge‑augmented belief networks, embed logical rules into continuous representations, enabling joint optimization over data and knowledge while retaining explicit predicates and rules for explanation. 


Any of the above lines of inquiry should ultimately lead to more sophisticated, efficient, and robust maintenance strategies, with the long-term objective of achieving the operational efficiency of fully autonomous maintenance. Here, the emerging paradigm is Agentic AI: systems that do not just predict failures, but set goals, plan multi-step actions, call tools, coordinate with other systems, and adapt over time.  Within this paradigm, the most promising trend is toward NESY agents, because they can combine neural perception and prediction with explicit reasoning, path finding, constraint handling, and explanation. \cite{Nisa_Shirazi_Saip_Pozi_2025} identify NESY as a key aspect of Agentic AI for reasoning and decision making.

In a maintenance context, an agentic system might continuously monitor sensor data, infer health states, diagnose root causes, propose and schedule interventions, order spare parts, and coordinate with operators or robots, all with minimal human oversight. NESY is well suited to this because many of these tasks are inherently symbolic: reasoning over maintenance procedures and safety rules, checking regulatory constraints, navigating plant layouts, following standard operating procedures, and updating work orders and logs. Neural components excel at perception and forecasting (e.g., RUL prediction from time series, anomaly detection, condition classification), while symbolic components capture domain knowledge as rules, ontologies, temporal logic constraints, and planning operators.

For example, a NESY maintenance agent could: (i) use a neural model to estimate the probability of a bearing failure within the next 200 hours; (ii) apply symbolic rules and a Bayesian or logical reasoner to perform root cause analysis across multiple components; (iii) invoke a planner that respects constraints expressed in temporal or signal temporal logic (e.g., \textit{do not shut down both redundant pumps simultaneously}, \textit{inspection must precede replacement within 24 hours}); and (iv) generate a human-readable maintenance plan linked to the underlying rules and sensor evidence. Because the reasoning layer is explicit, the agent’s decisions are auditable and can be checked against safety, regulatory, and contractual requirements.

Moreover, NESY agents would be able to integrate with domain specific ontologies and knowledge graphs, encoding assets, subsystems, failure modes, and work processes as graph-structured knowledge. This enables path finding and structured decision-making at the system level: which sequence of actions minimizes downtime given current inventory and staffing, which subsystems must be isolated before intervention, how failures propagate across a network of components. By grounding neural predictions in symbolic structure, Agentic NESY systems promise maintenance that is not only autonomous, but also robust, constrained by domain knowledge, and capable of providing justifications that engineers and regulators can trust.

\section{Conclusion}\label{sec:conclusion}
The systematic review of the State-of-the-art in Predictive Maintenance (PdM) over the last five years confirms that maintenance is transforming from a traditional reactive, ``fail-and-fix,'' philosophy into a critical strategic function and proactive discipline, driven by the technological foundation of Industry 4.0. We observed that PdM modeling approaches generally fall into three categories: 1) physics-based, 2) knowledge-based, and 3) data-driven, with the vast majority belonging to the third. While data-driven methods, particularly those leveraging deep learning, demonstrate high predictive accuracy, they suffer from significant limitations, including the dependence on large labeled datasets, poor out-of-distribution generalization, and a fundamental lack of transparency (``black-box'' nature). Conversely, traditional knowledge-based systems, while offering high explainability and the ability to incorporate expert domain knowledge, often struggle with poor accuracy and require ongoing expert supervision. Consequently, a clear trend toward hybrid and multi-model solutions can be seen across the spectrum of PdM applications.

This survey proposed moving beyond current hybrid systems by adopting Neuro-symbolic AI (NESY), which entails the tight integration of deep learning with symbolic logic. The research opportunities highlighted in this work, including the exploration of STL for time-series constraints and the joint modeling of multi-component systems using graph-based NESY structures, constitute concrete steps toward addressing these gaps. Moreover, overcoming persistent challenges, such as the scarcity of labeled run-to-failure data, the need for robust generalization, and the difficulty of fusing heterogeneous data, hinges on the capabilities offered by NESY. 

Ultimately, NESY serves as the central enabling technology for the next frontier in industrial maintenance: fully autonomous \textit{Agentic} systems. By combining the predictive power of neural networks with the explicit reasoning and constraint handling of symbolic logic, Agentic NESY systems can not only predict failures and estimate Remaining Useful Life (RUL) but also diagnose root causes, generate auditable maintenance plans, and coordinate complex actions within the operational workflow. This paradigm shift promises maintenance that is not only automated but also robust, trusted, and fully aligned with established domain knowledge.

\section{Limitations}
The volume of literature in the PdM domain exceeds what a single researcher can feasibly review in detail. Consequently, we relied extensively on natural language processing techniques to categorize and summarize thousands of studies into the most prevalent modeling approaches. While this strategy may have led to the omission of some relevant contributions addressing the identified gaps, our objective was to obtain a macro-level view of trends in the field and to identify opportunities for neuro-symbolic integration across approaches. We intend the ideas presented here to serve as a practical starting point for researchers from specialized subfields who seek to draw inspiration from adjacent areas, as well as seasoned NESY researchers interested in extending their work into PdM, and more broadly, time-series applications.

\newpage
\appendix

\section{Methodology}\label{app:methodology}
This review's methodology was guided by the principles proposed in \cite{Kitchenham07guidelinesfor}. 

\subsection{Research Questions}\label{sec:research_questions}
\begin{enumerate}
    \item What are the current State-of-the-art approaches for predictive maintenance in industrial settings?
    \item What are the limitations of current State-of-the-art approaches?
    \item How can the limitations of current approaches be addressed with neuro-symbolic techniques?
\end{enumerate}

\subsection{Search process}
\begin{figure}[ht]
\centering
\begin{subfigure}{0.45\textwidth}
   \includegraphics[width=1\linewidth]{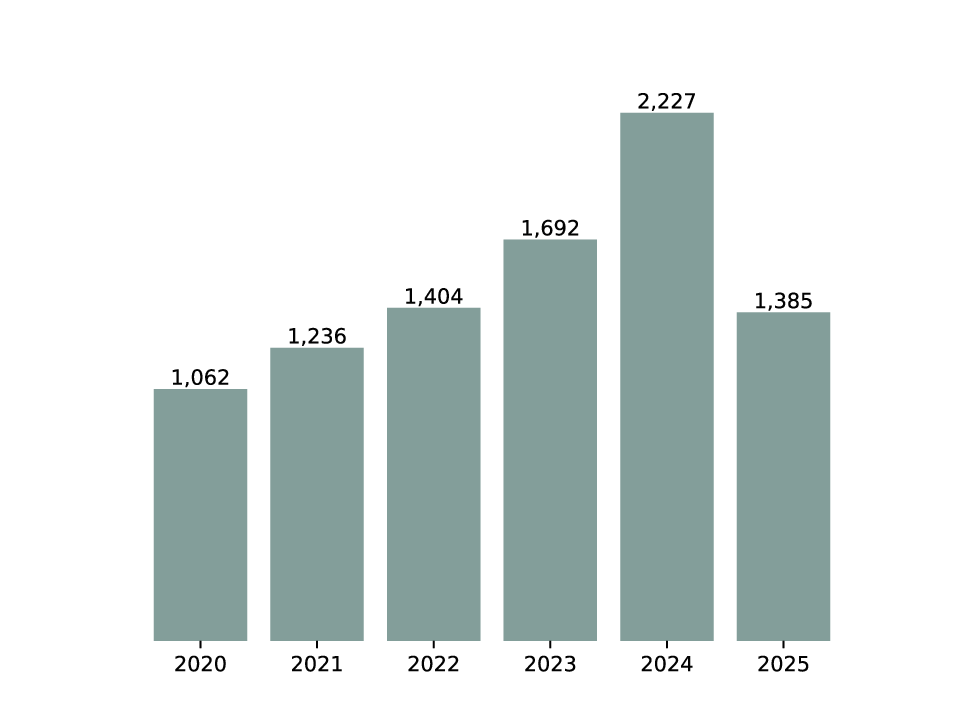}
    \caption{Number of publications per year. Note that the last date in 2025 is August 1st.}
    \label{fig:num_pub}
\end{subfigure}
\hfill
\begin{subfigure}{0.5\textwidth}
   \centering
    \includegraphics[width=1\linewidth]{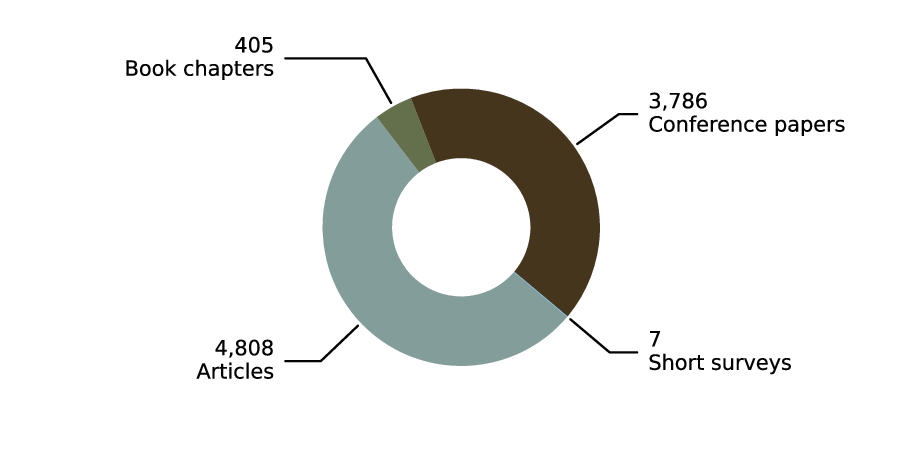}
    \caption{Document types returned by initial search}
    \label{fig:doc_types}
\end{subfigure}
\caption{Publications in the last five years}
\label{fig:num_publications_group}
\end{figure}

Scopus was chosen to perform the initial search, as Scopus indexes many top peer-reviewed journals and conference proceedings. It is possible some relevant studies were missed, but as the aim is to shed light on the field generally, the assumption is that these journals and proceedings are a representative body of research about the field. The query, provided in Table \ref{tab:scopus_query}), was restricted to the English language and the last 5 years, and produced a total of 9,005 results. Review articles were excluded. The results are made up of 4,808 journal articles, 3,786 conference papers, 405 book chapters, and 7 short surveys, as illustrated in Figure \ref{fig:doc_types}. An upward trend can be seen in the number of publications each year indicating a rising interest in the topic - Figure \ref{fig:num_pub}.

\begin{table}[ht]
    \centering
    \tiny
    \begin{tabular}{l}
(  \\
TITLE-ABS-KEY ( predictive maintenance ) OR  \\
TITLE-ABS-KEY ( PdM ) OR  \\
TITLE-ABS-KEY ( fault detection ) OR  \\
TITLE-ABS-KEY ( remaining useful life ) OR  \\
TITLE-ABS-KEY ( RUL )  \\
) \\
AND  \\
(  \\
ABS ( sensor ) OR  \\
ABS ( IoT )  \\
)  \\
AND  \\
PUBYEAR $>$ 2019 AND PUBYEAR $<$ 2026  \\
AND  \\
(  \\
EXCLUDE ( DOCTYPE , ``re'' ) OR  \\
EXCLUDE ( DOCTYPE , ``cr'' ) OR  \\
EXCLUDE ( DOCTYPE , ``dp'' ) OR  \\
EXCLUDE ( DOCTYPE , ``bk'' ) OR  \\
EXCLUDE ( DOCTYPE , ``tb'' ) OR  \\
EXCLUDE ( DOCTYPE , ``le'' ) OR  \\
EXCLUDE ( DOCTYPE , ``ed'' ) OR  \\
EXCLUDE ( DOCTYPE , ``er'' )  \\
)  \\
AND  \\
(  \\
LIMIT-TO ( SUBJAREA , ``ENGI'' ) OR  \\
LIMIT-TO ( SUBJAREA , ``COMP'' )  \\
)  \\
AND  \\
(  \\
LIMIT-TO ( LANGUAGE , ``English'' )  \\
)  \\
AND  \\
(  \\
EXCLUDE ( EXACTKEYWORD , ``Digital Storage'' ) OR  \\
EXCLUDE ( EXACTKEYWORD , ``Network Security'' ) OR  \\
EXCLUDE ( EXACTKEYWORD , ``Timing Circuits'' ) OR  \\
EXCLUDE ( EXACTKEYWORD , ``Electric Fault Currents'' ) OR  \\
EXCLUDE ( EXACTKEYWORD , ``Aircraft Detection'' ) OR  \\
EXCLUDE ( EXACTKEYWORD , ``Wireless Sensor Networks'' ) OR  \\
EXCLUDE ( EXACTKEYWORD , ``\&apos;current'' ) OR  \\
EXCLUDE ( EXACTKEYWORD , ``Wind Turbines'' ) OR  \\
EXCLUDE ( EXACTKEYWORD , ``Edge Computing'' ) OR  \\
EXCLUDE ( EXACTKEYWORD , ``Railroads'' ) OR  \\
EXCLUDE ( EXACTKEYWORD , ``Matlab'' ) OR  \\
EXCLUDE ( EXACTKEYWORD , ``Unmanned Aerial Vehicles (uav)'' ) OR  \\
EXCLUDE ( EXACTKEYWORD , ``Antennas'' ) OR  \\
EXCLUDE ( EXACTKEYWORD , ``Smart Power Grids'' ) OR  \\
EXCLUDE ( EXACTKEYWORD , ``Railroad Transportation'' ) OR  \\
EXCLUDE ( EXACTKEYWORD , ``Electric Inverters'' ) OR  \\
EXCLUDE ( EXACTKEYWORD , ``Electric Power Transmission Networks'' )  \\
)   \\
    \end{tabular}
    \caption{The Scopus index was queried on August 1st, 2025}
    \label{tab:scopus_query}
\end{table}

\subsection{Study selection process}\label{sec:study_selec_proc}
We limit the Scopus results by analyzing the titles and abstracts through the use of a Large Language Model (LLM), namely ``gpt-4.1-nano'' from OpenAI.\endnote{\url{https://openai.com/}} For each entry, we combine the title and abstract, and prompt the LLM to extract the following features:

\begin{enumerate}
    \item Domain 
    \item Input 
    \item Output 
    \item Modeling category 
    \item Modeling technique 
    \item Explainability 
    \item Relevance score 
\end{enumerate}

\begin{figure}
    \centering
    \includegraphics[width=0.9\linewidth]{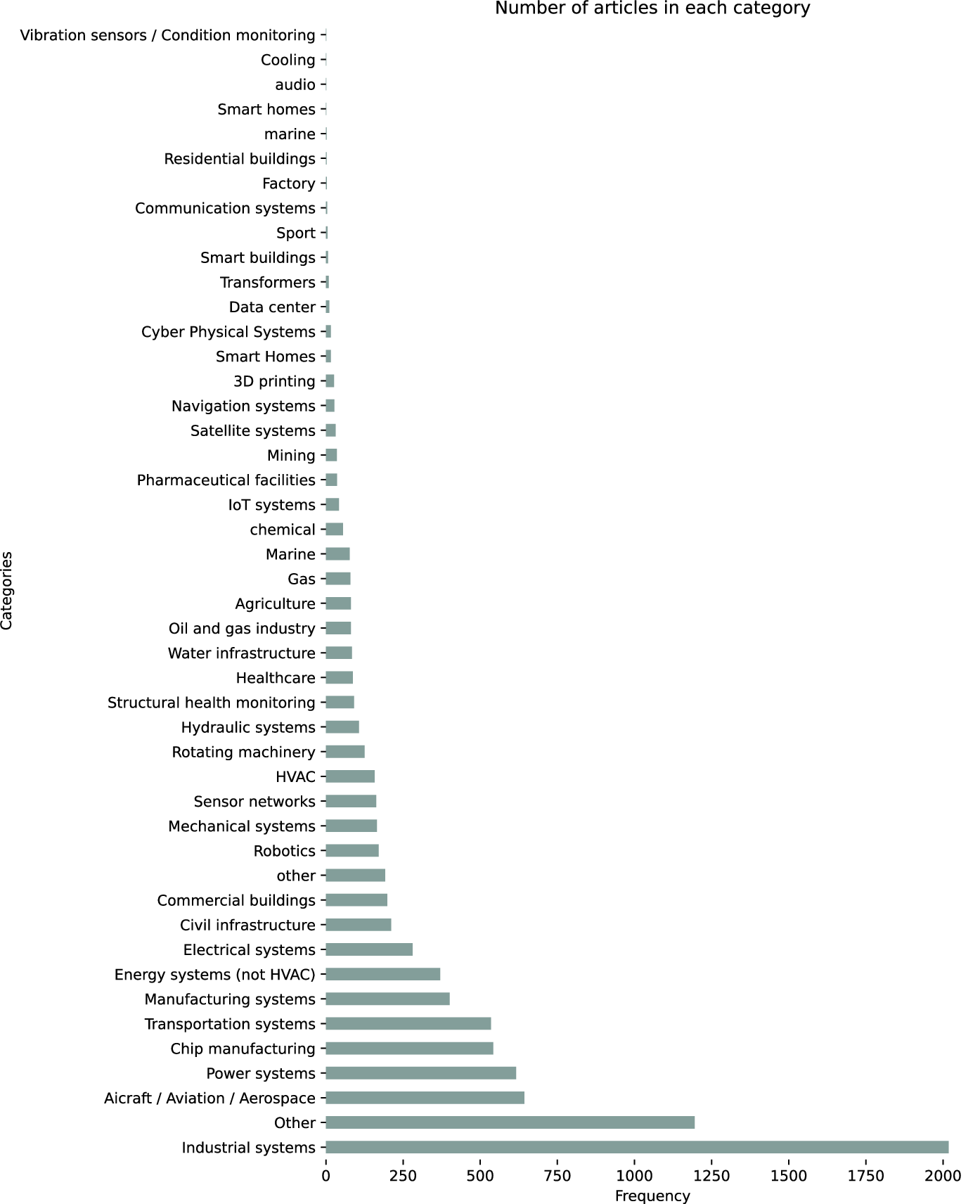}
    \caption{Number of articles in each domain.}
    \label{fig:domains}
\end{figure}

\begin{figure}
    \centering
    \includegraphics[width=.75\linewidth]{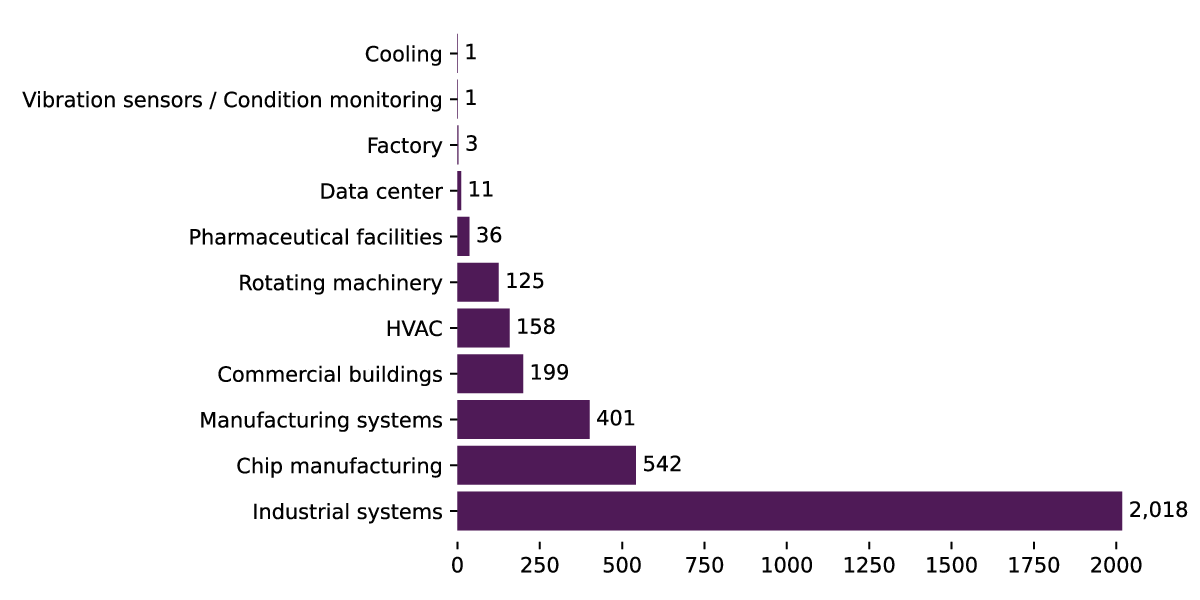}
    \caption{Number of articles in each of the filtered domains.}
    \label{fig:domain_filtered}
\end{figure}

The complete prompt is provided in Box \ref{box:prompt}. The results produced by the LLM contained inconsistencies in the naming of features and formatting. In particular, the Domain feature contained 4,273 unique values which is not practically useful. To address this issue, two new features were added: \textit{Domain category} and \textit{Domain subcategory}. The \textit{Domain} text strings were manually clustered into these new categories. Studies with similar category strings were consolidated under a unified label with the aid of OpenRefine.\endnote{\url{https://openrefine.org/}}. A \textit{Domain subcategory} was added manually where appropriate. This effort generated 46 unique domain categories (shown in Figure \ref{fig:domains}) which were used in the next exclusion step. Articles where the domain was not relevant (eg. \textit{3D Printing}, \textit{Sport}, \textit{Audio}, etc.) to our domains of interest were excluded. This reduced the number of studies to 3,495. The remaining domain categories include: \textit{Industrial systems}, \textit{Chip manufacturing}, \textit{Manufacturing systems}, \textit{Commercial buildings}, \textit{Rotating machinery}, \textit{Factory}, \textit{Vibration sensors / Condition monitoring}, \textit{Data center}, \textit{Pharmaceutical facilities}, \textit{HVAC}, and \textit{Cooling}. The number of articles in each domain category is listed in Figure \ref{fig:domain_filtered}.

\vspace*{10pt}

\begin{tcolorbox}[colframe=gray, colback=white, coltitle=white, sharp corners, title=Data Extraction Prompt] \label{box:prompt}

\footnotesize
PROMPT = \verb|'''|\\
You are an expert in predictive maintenance for industrial systems. Given the following abstract or summary of a research paper, classify it according to these fields:

\begin{enumerate}
    \item Domain: examples of domain include but are not limited to: ``HVAC systems'', ``Commercial buildings'', ``Pharmaceutical facilities'', ``Chip manufacturing''. 
    \item Input: types of input data including but not limited to: ``Sensor data'', ``Time series'', ``Natural language text'', ``Structured data'', ``Unstructured data'', ``Images''.
    \item  Output: type of predicted output including but not limited to: ``Remaining Useful Life'', ``Fault prediction'', ``Root cause analysis''.
    \item Modeling category: one of [``Rules and/or logic based'', ``Physics based'', ``First principles'', ``Knowledge based'', ``Data driven'', ``Hybrid'', ``Neurosymbolic'', ``Physics informed'', ``Other''].
    \item Modeling technique: specific techniques mentioned (e.g., ``SVM'', ``GNN'', ``CNN'', ``PINN'', etc.).
    \item Explainability: whether it uses interpretable or explainable techniques (``Yes: <technique>'' or ``No'').
    \item Relevance score: integer from 0 (lowest) to 5 (highest), where relevance is defined as the study's applicability to predictive maintenance in HVAC systems for commercial buildings, pharmaceutical facilities, or chip manufacturing, using sensor data, combining rules and ML, and being interpretable.
\end{enumerate}

When your answer is 'Other', think again. Analyse the text to see if you can figure out the domain.\\
Provide your answers in JSON format where the keys are the field names from the above list.\\
Abstract: \verb|"""{text}"""|\\
Answer:\\
\verb|'''|

\end{tcolorbox}

\subsection{Data synthesis}

\begin{figure}[ht]
\centering
\begin{subfigure}{0.49\textwidth}
    \centering
    \includegraphics[width=1\linewidth]{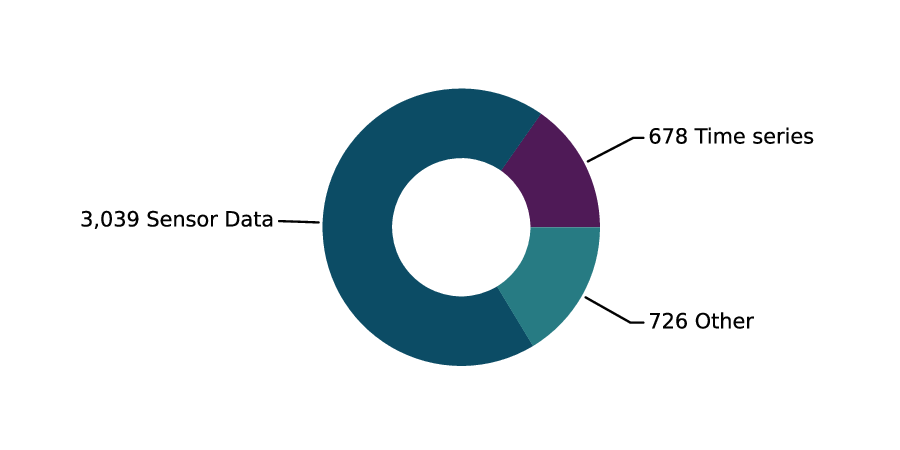}
    \caption{Number of articles in each of the input categories.}
    \label{fig:Input_categories}
\end{subfigure}
\begin{subfigure}{0.49\textwidth}
    \centering
    \includegraphics[width=1\linewidth]{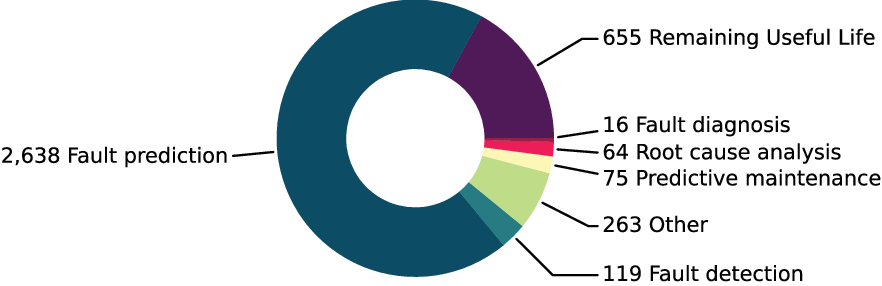}
    \caption{Number of articles in each of the output categories.}
    \label{fig:Output_categories}

\end{subfigure}
\caption{Input and Output categories.}
\label{fig:input_ouput_group}
\end{figure}

\begin{figure}[ht]
\begin{subfigure}{0.49\textwidth}
    \centering
    \includegraphics[width=1\linewidth]{figures/Modeling_categories.eps}
    \caption{Number of articles in each of the modeling categories.}
    \label{fig:Modeling_categories}
\end{subfigure}
\begin{subfigure}{0.49\textwidth}
    \centering
    \includegraphics[width=1\linewidth]{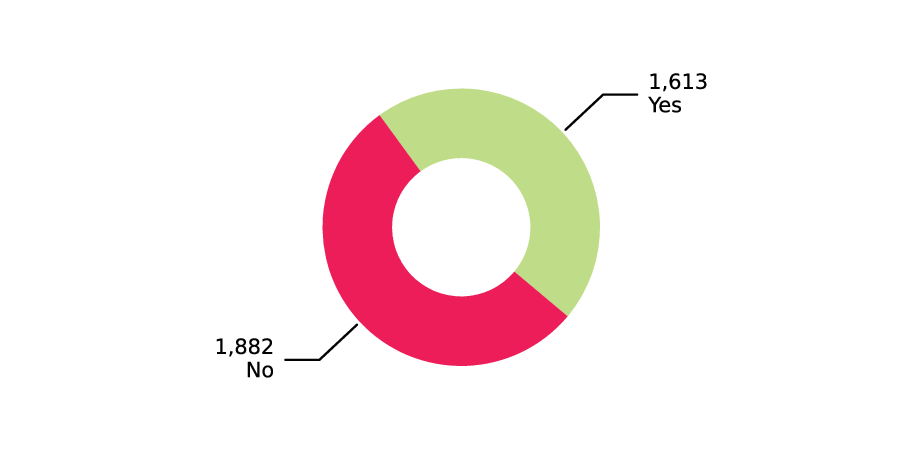}
    \caption{Number of articles where the solution is said to be explainable.}
    \label{fig:explainability}
\end{subfigure}
\caption{Modeling categories and Explainability.}
\label{fig:mod_explain_group}
\end{figure}

\begin{figure}
    \centering
    \includegraphics[width=1\linewidth]{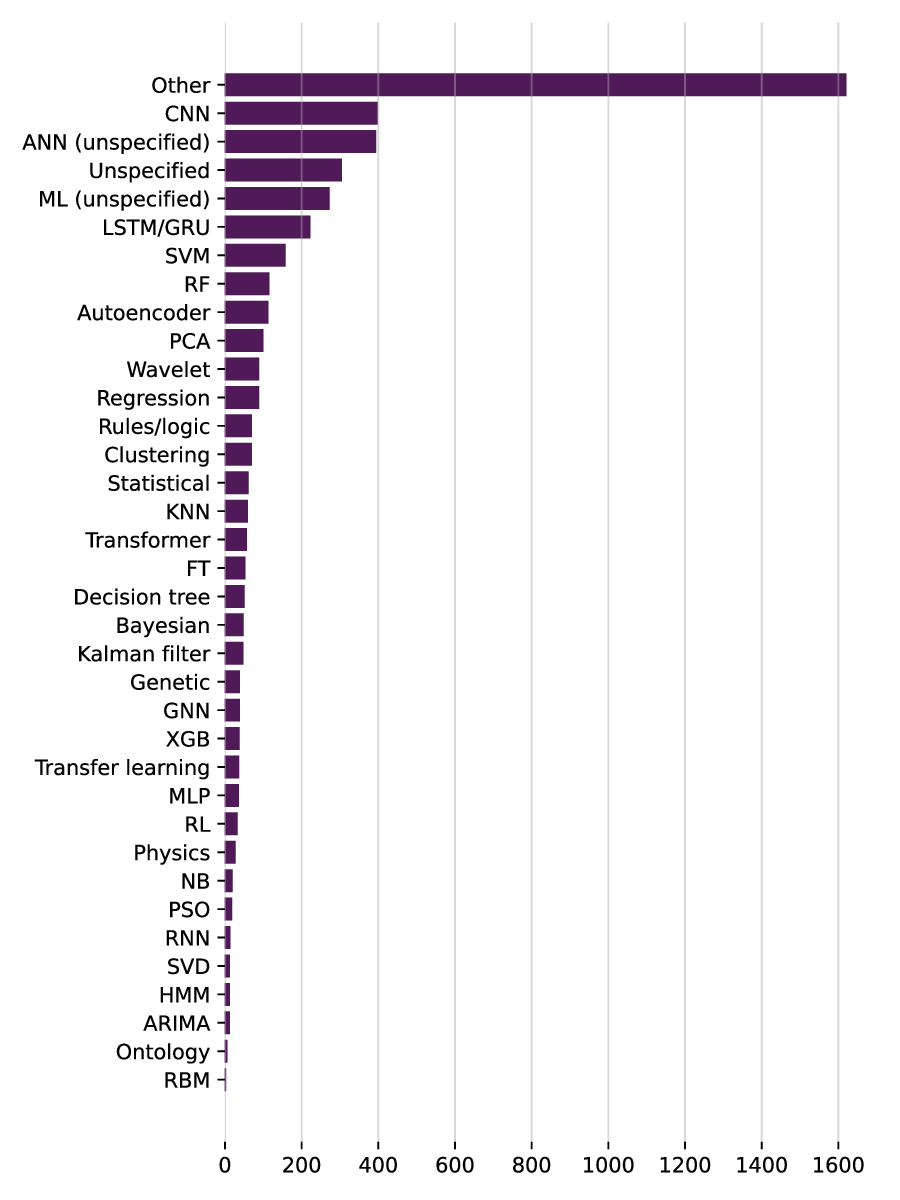}
    \caption{Number of articles for each of the modeling techniques.}
    \label{fig:Mod_tech_counts}
\end{figure}

This section aims to concisely answer Research Question 2: \textit{What are the current State-of-the-art approaches for predictive maintenance in industrial settings?} For each of the extracted features, we count the number of articles for each value of the given feature. These are provided in Figures \ref{fig:Input_categories}, \ref{fig:Output_categories}, \ref{fig:Modeling_categories}, \ref{fig:explainability}, and \ref{fig:Mod_tech_counts}. It is clear that the majority of research is dedicated to data-driven approaches. The use of Artificial Neural Networks (ANNs) is particularly prevalent, followed by Principal Component Analysis (PCA).

\newpage
\section{Publicly available datasets} \label{app:datasets}
\begin{landscape}
\begin{table}[ht]
    \centering
    \footnotesize
    \renewcommand{\arraystretch}{1.5}
    \begin{tabular}[b]{%
    >{\raggedright\arraybackslash}p{2cm}%
    >{\raggedright\arraybackslash}p{7cm}%
    >{\raggedright\arraybackslash}p{8cm}}
    \toprule
        Ref. &  Link & Description \\
        \midrule
        \cite{OEDI_Dataset_5763} & \url{https://data.openei.org/submissions/5763} & These datasets can be used to evaluate and benchmark the performance accuracy of Fault Detection and Diagnostics (FDD) algorithms or tools. It contains operational data from simulation, laboratory experiments, and field measurements from real buildings for seven HVAC systems/equipment (rooftop unit, single-duct air handler unit, dual-duct air handler unit, variable air volume box, fan coil unit, chiller plant, and boiler plant). \\
        \cite{Saxena_2008} & \url{https://ti.arc.nasa.gov/tech/dash/groups/pcoe/prognostic-data-repository/} & NASA Turbofan engine degradation dataset\\
        \cite{Mauthe_Hagmeyer_Zeiler_2021} & \url{https://doi.org/10.36001/ijphm.2021.v12i2.3087} & Creation of Publicly Available Data Sets for Prognostics and Diagnostics Addressing Data Scenarios Relevant to Industrial Applications. See appendix for links to individual datasets\\
        \cite{Granderson2020} & \url{https://figshare.com/articles/dataset/LBNLDataSynthesisInventory_pdf/11752740/3}. &  Dataset for building fault detection and diagnostics algorithm creation and performance testing. \\
        \cite{NASA_femto} &  & FEMTO bearing dataset  - Experiments on bearings’ accelerated life tests provided by FEMTO-ST Institute, Besançon, France.\\
        Prognostic Challenge Datasets & \url{https://phmsociety.org/?s=data+challenge} & PHM Society-Conference: These conferences often organize prognostic challenges that provide valuable datasets and competition opportunities. \\
        NASA Ames Prognostics Data Repository (PCoE) &  \url{https://www.nasa.gov/intelligent-systems-division/discovery-and-systems-health/pcoe/pcoe-data-set-repository/}, \url{https://data.phmsociety.org/nasa/} & This repository collects many datasets specifically generated to demonstrate prognostic approaches.\\
        
      \bottomrule  
    \end{tabular}
    \caption{Publicly available data sets.}
    \label{tab:public_datasets}
\end{table}
\end{landscape}

\newpage
\section{Ontology and Bayesian Networks Examples}\label{app:ont-bayes}
\begin{landscape}
\begin{table}[ht]
    \centering
    \footnotesize
    \renewcommand{\arraystretch}{1.5}
    \begin{tabular}[b]{%
    >{\raggedright\arraybackslash}p{2cm}%
    >{\raggedright\arraybackslash}p{4cm}%
    >{\raggedright\arraybackslash}p{4cm}%
    >{\raggedright\arraybackslash}p{7cm}}
    \toprule
        Scenario & Standards & Example Nodes & Diagnosis and Interventions \\
    \midrule
        Centrifugal pump: cavitation vs. misalignment & ISO 15926 for plant topology; QUDT for units; MIMOSA/IOF for failure modes and maintenance tasks & NPSH available, suction pressure, flow rate, temperature (vapor pressure), pump speed, cavitation, vibration band energy, seal wear, maintenance state (strainer cleanliness). & Given evidence (high vibration in specific bands, low suction pressure, elevated temperature), the CBN yields higher $P(cavitation)$ than $P(misalignment)$, explaining propagation to seal wear. Interventions: $do(reduce \ speed)$ or $do(clean \ strainer)$ are simulated; model estimates best action to reduce cavitation risk and downtime, and the ontology links the recommended task to parts and work instructions. \\

        Wind turbine: gearbox bearing wear & SSN/SOSA for SCADA and CMS vibration sensors; IOF/MIMOSA for drivetrain taxonomy and failure modes; QUDT for units; site/environment context (wind, turbulence). & Wind speed/turbulence, rotor speed, torque, oil temperature, particle count, vibration features (BPFO/BPFI bands), bearing wear, remaining useful life (RUL). & Evidence of rising spectral energy at BPFO and increasing particle count updates $P(bearing \ wear)$. Interventions: $do(curtailment \ under \ high \ turbulence)$ or $do(oil \ change)$ are evaluated for their effect on $P(failure \ within \ 90 \ days)$ and RUL; ontology resolves which kit and crane logistics apply to this turbine. \\
        
        Buildings HVAC: coil fouling (chiller/AHU) & Brick schema for HVAC equipment/points; SAREF/SSN for observations; maintenance ontology for tasks and costs. & Filter differential pressure, valve position, coil approach temperature, airflow, fouling state, energy consumption. & With elevated $\Delta P$ and worsening approach temperature, the CBN diagnoses fouling and estimates energy penalty. Interventions: $do(replace \ filters)$ vs. $do(schedule coil \ cleaning)$ are compared; the ontology populates a CMMS work order with required parts, skills, and estimated downtime. \\

        Rail rolling stock: wheel flats vs. bearing faults & Asset hierarchy $(car \rightarrow bogie \rightarrow axle \rightarrow wheel/bearing)$; sensor metadata (axle-box temp, vibration); inspection/work-order history & Braking events, wheel slip, impact forces, wheel flat, bearing temperature rise, bearing fault. & Given wayside impact detections and temperature trends, the CBN separates wheel flats (causing impact, secondary heating) from primary bearing defects and simulates $do(speed \ restriction)$ vs. $do(immediate \ inspection)$ to manage risk until depot access. \\
    \bottomrule
    \end{tabular}
    \caption{Examples of the combination of ontologies and BNs for diagnosis and intervention recommendations.}
    \label{tab:ontology-cbn}
\end{table}
\end{landscape}

\begin{acks}
This work is supported by Taighde Éireann – Research Ireland under Grant Number Research Ireland/12/RC/2289\_P.
\end{acks}

\theendnotes
\bibliographystyle{SageH}
\bibliography{references}

\end{document}